\documentclass[lettersize,journal]{IEEEtran}
\usepackage{amsmath,amsfonts}
\usepackage{algorithmic}
\usepackage{algorithm}
\usepackage{array}
\usepackage[caption=false,font=normalsize,labelfont=sf,textfont=sf]{subfig}
\usepackage{textcomp}
\usepackage{stfloats}
\usepackage{url}
\usepackage{verbatim}
\usepackage{graphicx}
\usepackage{cite}
\usepackage{multirow}
\usepackage[pagebackref=true,breaklinks=true,letterpaper=true,colorlinks,citecolor=blue,linkcolor=blue,bookmarks=false]{hyperref}
\begin{document}

\title{Limb-Aware Virtual Try-On Network with Progressive Clothing Warping}

\author{Shengping Zhang, Xiaoyu Han, Weigang Zhang, Xiangyuan Lan, Hongxun Yao, Qingming Huang
\thanks{
Manuscript received 3 September 2022; revised 11 April 2023 and 28 May 2023; accepted 5 June 2023. Date of publication 14 June 2023; date of current version 23 January 2024. \textit{(Corresponding author: Weigang Zhang.)}
}
        
\thanks{Shengping Zhang, Xiaoyu Han, and Weigang Zhang are with Harbin Institute of Technology, Weihai 264209, China (e-mail: s.zhang@hit.edu.cn; xyhan@stu.hit.edu.cn; wgzhang@hit.edu.cn).}
\thanks{Xiangyuan Lan is with the Peng Cheng Laboratory, Shenzhen 518055, China (e-mail: lanxy@pcl.ac.cn).}
\thanks{Hongxun Yao is with Harbin Institute of Technology, Harbin 150001, China (e-mail: h.yao@hit.edu.cn).}
\thanks{Qingming Huang is with the University of Chinese Academy of Sciences, Beijing 100190, China (e-mail: qmhuang@ucas.ac.cn).}
\thanks{
The code is available at \url{https://github.com/aipixel/PL-VTONv2}.
}
}

\maketitle
\begin{abstract}
Image-based virtual try-on aims to transfer an in-shop clothing image to a person image.
Most existing methods adopt a single global deformation to perform clothing warping directly, which lacks fine-grained modeling of in-shop clothing and leads to distorted clothing appearance.
In addition, existing methods usually fail to generate limb details well because they are limited by the used clothing-agnostic person representation without referring to the limb textures of the person image. 
To address these problems, we propose Limb-aware Virtual Try-on Network named PL-VTON, which performs fine-grained clothing warping progressively and generates high-quality try-on results with realistic limb details.
Specifically, we present Progressive Clothing Warping (PCW) that explicitly models the location and size of in-shop clothing and utilizes a two-stage alignment strategy to progressively align the in-shop clothing with the human body. Moreover, a novel gravity-aware loss that considers the fit of the person wearing clothing is adopted to better handle the clothing edges.
Then, we design Person Parsing Estimator (PPE) with a non-limb target parsing map to semantically divide the person into various regions, which provides structural constraints on the human body and therefore alleviates texture bleeding between clothing and body regions.
Finally, we introduce Limb-aware Texture Fusion (LTF) that focuses on generating realistic details in limb regions, where a coarse try-on result is first generated by fusing the warped clothing image with the person image, then limb textures are further fused with the coarse result under limb-aware guidance to refine limb details.
Extensive experiments demonstrate that our PL-VTON outperforms the state-of-the-art methods both qualitatively and quantitatively.
\end{abstract}


\section{Introduction}
\IEEEPARstart{V}{irtual} try-on refers to an image synthesis task of transferring an in-shop clothing image to a person image while preserving characteristics in other regions, which has generated significant research interest due to the rise of online shopping. 
Existing virtual try-on methods can be divided into 3D-based methods~\cite{guan2012drape, pons2017clothcap, chen2016synthesizing, sekhavat2016privacy} and image-based methods~\cite{viton, cp-vton, han2019clothflow, jandial2020sievenet, jetchev2017conditional, zflow, yang2021ct, ovnet, hu2022spg, liu2019swapgan, xu2021virtual, kim2019style, zflow, vtnfp, wuton, viton-gt, dcton, ovnet, viton-hd, viton-gt2, flow-with-style, sdafn, dresscode, hr-vton} according to the type of processed data. 
3D-based methods utilize computer graphics for building 3D models and obtain try-on results through complex rendering. Although these methods have excellent control over clothing material and deformation, they rely heavily on complex data representations and consume massive computational resources. Conversely, image-based methods have a wider range of applications due to lightweight data.

\begin{figure}[!t]
  \resizebox{\linewidth}{!} {
    \includegraphics{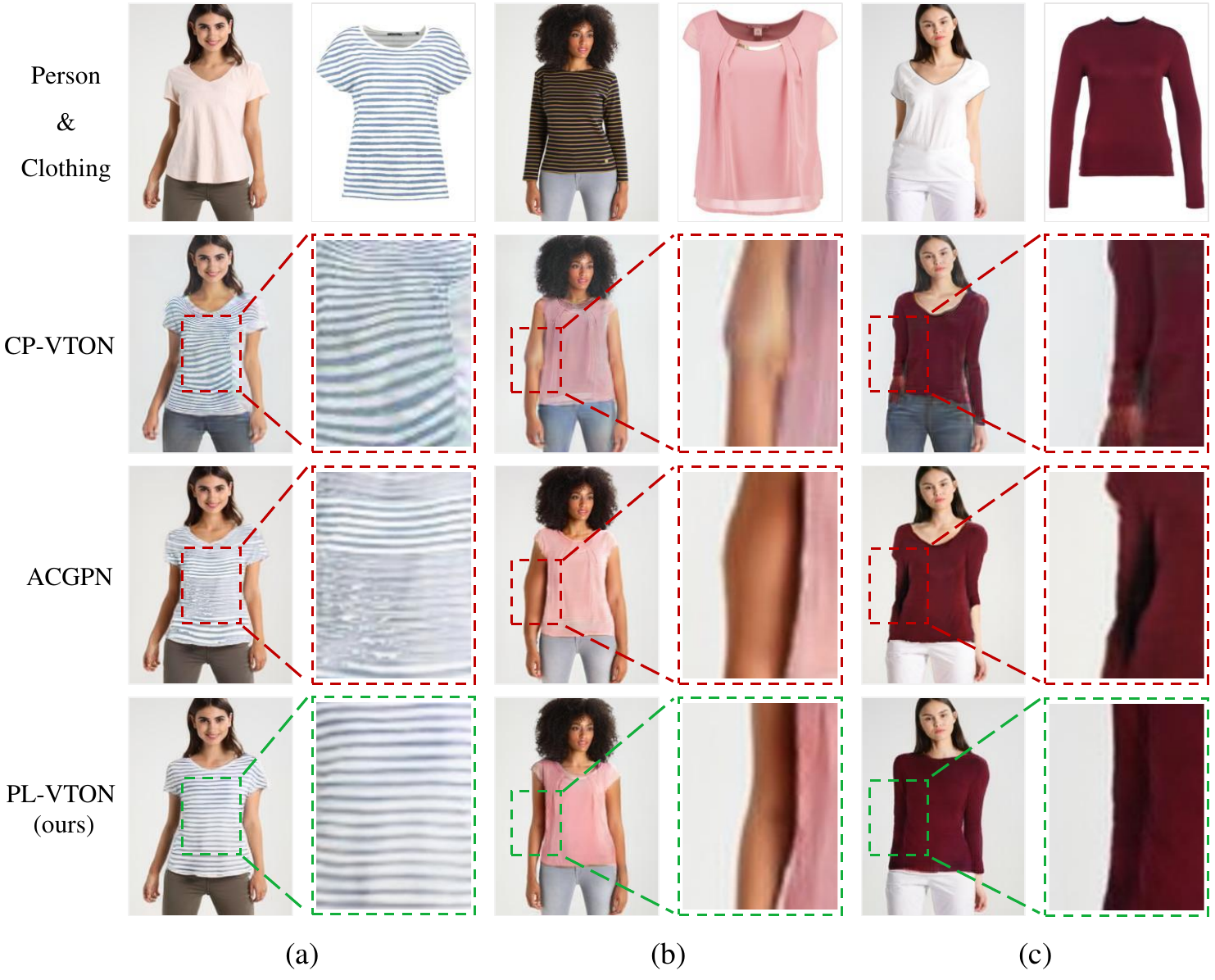}
  }
  \caption{Fine-grained Clothing warping and realistic limb detail generation are challenging in image-based virtual try-on, especially when transforming between long-sleeved and short-sleeved clothing. Compared with existing methods, our proposed method achieves superior performance in these cases.}
  \label{fig:teaser}
\end{figure}

\begin{figure}[!t]
    \centering
    \resizebox{\linewidth}{!} {
        \includegraphics{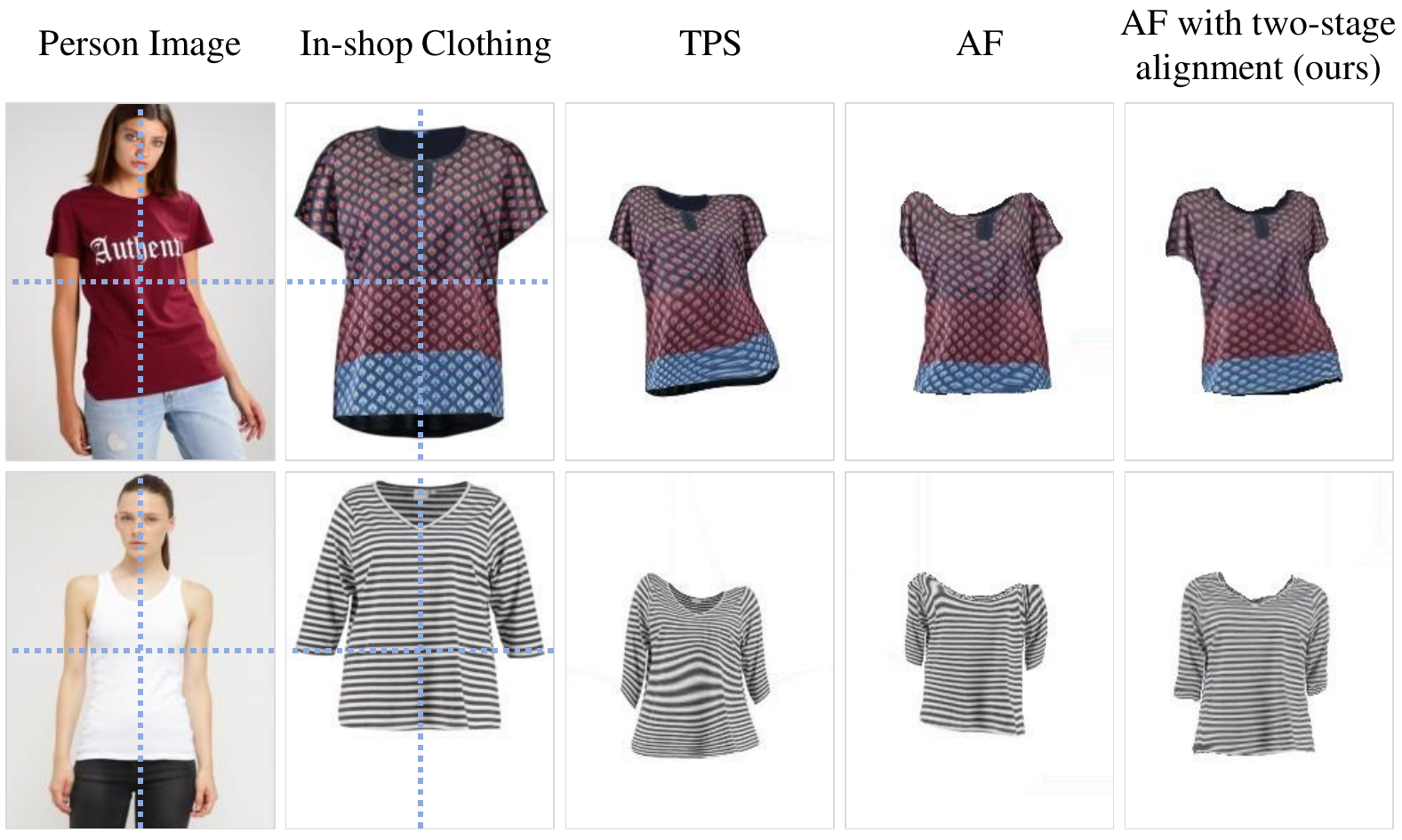}
    }
    \caption{The finesse and stability of different clothing warping strategies, where TPS represents the warped results based on thin-plate spline transformation, AF represents the warped results based on an appearance flow, and the dotted lines are drawn to show the inconsistency in location and size of the person's original clothing and the in-shop clothing more intuitively.}
    \label{fig:TPS}
\end{figure}

However, it is difficult to simulate the realistic effect of clothing warping in the plane dimension due to fewer semantic representations, which is a major challenge faced by previous image-based virtual try-on methods.
VITON~\cite{viton} first proposes the idea of using thin-plate spline (TPS) transformation to describe clothing warping, which is improved by subsequent CP-VTON~\cite{cp-vton}, ACGPN~\cite{acgpn}, etc.
Nonetheless, limited degrees of freedom require each parameter of TPS to control a large area of pixels, which leads to the warped result with inadequate details in the case of complex human poses.
Some other methods~\cite{han2019clothflow, pf-afn} use an appearance flow to perform clothing warping. However, due to insufficient controllability, the process of estimating pixel-wise displacements usually causes unstable and drastic results when dealing with detailed textures.
The above methods both regard the clothing warping as a single global deformation and adopt the neural network to predict the warped result directly, which does not explicitly consider the inconsistency in location and size of the person's original clothing and the in-shop clothing.
In this case, the warping network is required to allocate additional attention to implicitly translating and scaling the in-shop clothing image, which is a black-box warping process and has lower finesse and stability (as shown in Fig. \ref{fig:teaser} (a) and Fig. \ref{fig:TPS}). 

Another challenge of image-based virtual try-on is to generate realistic details in limb regions of the try-on result, especially in the case of transformation between long-sleeved and short-sleeved clothing, which is shown in Fig. \ref{fig:teaser} (b) and (c). Most existing methods occlude a wide range of source textures in the person image for a clothing-agnostic person representation to eliminate the effect of original clothing so that the network can be trained in a self-supervised way.
Some of them~\cite{mpv, chen2021fashionmirror} try to occlude clothing regions of the person image along clothing edges. However, the rough outline of the clothing is still reflected through non-clothing regions after the edgewise occlusion, which does not keep clothing-agnostic strictly and is negative for self-supervised training. 
Others~\cite{cp-vton, minar2020cp, viton, jandial2020sievenet} occlude both clothing regions and limb regions, which completely loses the reference of the limb textures in the person image and is not conducive to the limb detail generation.
Generally, limb textures of the person image are necessary as auxiliary information to generate the limb details in the try-on result, but they are limited by the clothing-agnostic person representation at the same time, it is difficult to find a balanced solution to take advantage of limb textures in inputs while eliminating the impact of the person's original clothing on the try-on result.

To address the above problems, we propose Limb-aware Virtual Try-on Network (PL-VTON) to perform more fine-grained and controllable clothing warping progressively and generate realistic limb details in the try-on result with limb-aware guidance, which consists of three sub-modules: Progressive Clothing Warping (PCW), Person Parsing Estimator (PPE), and Limb-aware Texture Fusion (LTF). 
Specifically, PCW adopts a two-stage alignment strategy that first explicitly models the location and size of the in-shop clothing through a pre-alignment network, then a multi-scale flow predictor is used to estimate the pixel-level geometric deformation of the clothing. In addition, considering the fit of the person wearing clothing in real scenarios, we also introduce a novel gravity-aware loss to better handle the clothing edges and make the warped result more realistic. 
Next, we adopt PPE to provide structural constraints on the human body for the subsequent limb texture extraction and generation of the try-on result by predicting a target parsing map of the person wearing the target clothing, where a non-limb target parsing map that contains the semantic information of the warped clothing is adopted for a smooth transition of prediction. 
Finally, LTF performs a coarse-to-fine texture fusion process with limb-aware guidance, where a coarse try-on result is first generated by fusing textures between the warped clothing and the person image according to the target parsing map, then limb textures that hide the geometric information of the person's original clothing are further fused with the coarse result while keeping clothing-agnostic to get a fine try-on result, where realistic details in limb regions are generated.
Moreover, considering the negative impact of the inappropriate occlusion in inputs, we design an improved clothing-agnostic person representation to better preserve the person's other characteristics except for clothing and limbs during try-on, where the occluded regions are semantically corrected by the person's parsing map.

The main contributions of this paper are as follows:
\begin{itemize}
\item
We propose Limb-aware Virtual Try-on Network (PL-VTON) that generates high-quality try-on results through three sub-modules with an improved clothing-agnostic person representation. 
\item
We design a novel Progressive Clothing Warping module that utilizes a two-stage alignment strategy to warp the in-shop clothing image progressively. Besides, a novel gravity-aware loss is adopted to better fit the warped clothing to the human body. 
\item
We present a novel Limb-aware Texture Fusion module to fuse textures of the warped clothing and the human body from coarse to fine, where limb-aware guidance has a positive impact on the reference to the limb textures of the person image, promoting the generation of limb details in the try-on result. 
\item
Extensive experiments on the VITON~\cite{viton} dataset demonstrate the superior performance of our PL-VTON compared to the state-of-the-art methods.
\end{itemize}

Our method is an extension of our previous work on the conference version~\cite{pl-vton}. Compared with \cite{pl-vton}, improvements made in this paper include: (1) In Progressive Clothing Warping, we adopt a new pre-alignment network to regress dynamic parameters of translation and scaling, which improves the robustness of parameter acquisition and avoids the negative impact of special human poses on the transformation parameters. (2) Considering the fit of the person wearing clothing in real scenarios, we propose a novel gravity-aware loss to better handle the clothing edges and make the warped result more realistic. (3) We introduce a non-limb target parsing map that contains the semantic information of the warped clothing in the middle process of Person Parsing Estimator, which serves as a geometric prior and assists in the smooth transition of predicting the target parsing map. (4) We further optimize the clothing-agnostic person representation by the semantic correction according to the person's parsing result to better retain the person's other characteristics except for clothing and limbs in the try-on result. (5) We conduct more qualitative and quantitative experiments. Specifically, we additionally introduce Structural Similarity (SSIM)~\cite{wang2004image} and Peak Signal to Noise Ratio (PSNR)~\cite{hore2010image} to further measure the quality of the generated try-on result. Then, we add three new ablation studies to evaluate each proposed contribution of our PL-VTON in more detail. Finally, additional try-on results of our method on the MPV~\cite{mpv}, VITON-HD~\cite{viton-hd}, and Dress Code~\cite{dresscode} datasets are further provided.

\section{Related Work}
\subsection{Flow Estimation}
Flow estimation describes a pixel-level correspondence that aims to find which pixels in the source can be used to synthesize the target and how those pixels are transferred to specific positions in the result.
Traditional flow estimation is usually used in video tasks, which is called optical flow estimation. There is a Siamese network that takes two consecutive video frames as inputs and generates a pixel-level field of the same size, which warps the raw pixels of the previous frame to the subsequent one.
FlowNet~\cite{dosovitskiy2015flownet} first proposes an end-to-end convolutional neural network to predict the optical flow directly. Besides, a large Flying Chairs dataset is synthesized for the optical flow estimation.
FlowNet2~\cite{ilg2017flownet} adopts a stacked hourglass network with more complex training strategies to improve the accuracy of optical flow estimation in small motion areas.
PWC-Net~\cite{sun2018pwc} predicts the optical flow using well-established principles of pyramidal processing, warping, and cost volume processing, which further promotes the performance and reduces the model size simultaneously.
RAFT~\cite{teed2020raft} proposes a novel deep network architecture with a recurrent unit, where many lightweight recurrent update operators are used to estimate the optical flow.

On the other hand, flow estimation is also applied to predict a 2D vector field without timing information, which warps the source image to the target based on the similarity in appearance. It is defined as the appearance flow by Zhou et al.~\cite{zhou2016view}.
The appearance flow has been widely applied in the field of computer vision.
For instance, StructureFlow~\cite{ren2019structureflow} adopts the appearance flow in image inpainting. To generate realistic alternative contents for missing holes, the offset vectors are predicted to flow the pixels from source regions to missing regions.
Additionally, document image rectification~\cite{ma2018docunet, li2019document, markovitz2020can, feng2021doctr} adopts an appearance flow network to regress a dense 2D vector field that samples pixels from a distorted document image to the rectified one.
The appearance flow is also employed in human pose transfer. Li et al.~\cite{li2019dense} fit a 3D model to a given pose and project it back to the 2D plane to compute the dense flow. With the pixel-level displacements, feature warping is performed on the person image, and the result with the target pose is generated.
In this paper, we adopt the appearance flow to describe clothing warping, which serves as the prior for generating a try-on result.

\begin{figure*}[!t]
    \centering
    \resizebox{\linewidth}{!} {
        \includegraphics{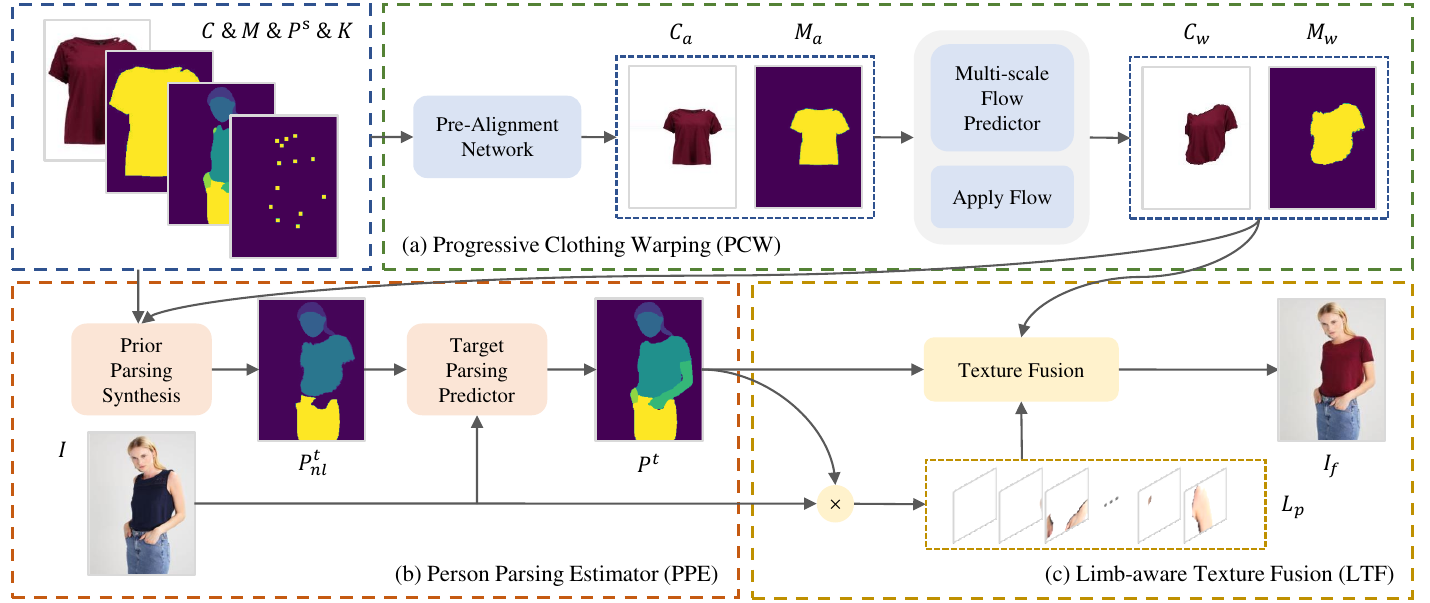}
    }
    \caption{The overview of the proposed PL-VTON. (a) PCW takes an in-shop clothing image $C$, an in-shop clothing mask $M$, a source parsing map $P^s$, and a keypoint map $K$ as inputs to generate a warped clothing image $C_w$ through a two-stage alignment strategy. (b) PPE first synthesizes a non-limb target parsing map $P^t_{nl}$ through a semantic combination and then uses this prior parsing result to predict a target parsing map $P^t$. (c) LTF performs a texture fusion process between the warped clothing and the person image based on a set of limb map patches $L_p$ and outputs the try-on result $I_f$.}
    \label{fig:framework}
\end{figure*}

\subsection{Fashion Analysis and Virtual Try-on}
Fashion analysis has attracted considerable attention in recent years due to its wide range of applications, whose initial emphasis is largely on the classical tasks such as clothing classification~\cite{chen2012describing}, clothing recognition~\cite{yang2011real}, clothing segmentation~\cite{yang2014clothing, wang2011blocks}, and fashion image retrieval~\cite{corbiere2017leveraging, lang2020plagiarism, d2021localized, chen2017query}. With the increasing diversification of computer vision, more novel tasks for fashion analysis such as fashion style estimation~\cite{al2020modeling, kiapour2014hipster, hsiao2017learning}, clothing recommendation~\cite{han2017learning, hidayati2018dress, hidayati2020dress, ding2021modeling}, popularity predicting~\cite{yamaguchi2014chic}, and fashion landmark localization~\cite{lee2019global, liu2016fashion, wang2018attentive} have been explored gradually, where virtual try-on is one of the most challenging tasks because it requires not only the accurate capture of clothing warping but also the reasonable texture fusion between the clothing and the human body.

Existing methods for virtual try-on are mainly based on 3D modeling~\cite{guan2012drape, pons2017clothcap, chen2016synthesizing, sekhavat2016privacy} or 2D images~\cite{viton, cp-vton, han2019clothflow, jandial2020sievenet, jetchev2017conditional, zflow, yang2021ct, ovnet, hu2022spg, liu2019swapgan, xu2021virtual, kim2019style, zflow, vtnfp, wuton, viton-gt, dcton, ovnet, viton-hd, viton-gt2, flow-with-style, sdafn, dresscode, hr-vton}.
3D-based methods rely on sophisticated instruments and higher-dimensional calculations to achieve the try-on effect, but the complicated process and data constrain the application value. Conversely, image-based methods are more suitable for real-world scenarios.
VITON~\cite{viton} first proposes to use the thin-plate spline (TPS) transformation to warp in-shop clothing, and then an encoder-decoder is adopted to generate the try-on result. Besides, it adopts a clothing-agnostic person representation to eliminate the effect of the original clothing item in the person image.
After that, CP-VTON~\cite{cp-vton} proposes to utilize a convolutional network to dynamically predict the TPS parameters, which further improves the accuracy of the clothing warping process.
CP-VTON+~\cite{minar2020cp} is designed based on the pipeline structure of CP-VTON, which corrects the erroneous clothing-agnostic person representation and makes the network input more reasonable.
WUTON~\cite{wuton} adopts the TPS transformation on the feature maps rather than at the pixel level.
ACGPN~\cite{acgpn} proposes an adaptive content generation and preservation scheme, which for the first time takes the semantic layout into consideration to generate try-on results.
DCTON~\cite{dcton} disentangles virtual try-on as clothing warping, skin synthesis, and image composition, where three modules are integrated within one framework for end-to-end cycle consistent training.
VITON-HD~\cite{viton-hd} and Morelli et al.~\cite{dresscode} propose to generate high-resolution try-on results with the TPS transformation.
However, the above TPS-based methods cannot handle a large geometric deformation well due to the limited degrees of freedom. To warp the in-shop clothing image more freely, some recent methods adopt the appearance flow in their warping strategies.
ClothFlow~\cite{han2019clothflow} first proposes a generative model based on the appearance flow to enhance the expressive ability of the clothing warping. By estimating the pixel-level flow between the source and target clothing regions, the model can effectively simulate geometric changes and get the warped result.
Zflow~\cite{zflow} adopts a gated aggregation of hierarchical flow estimates as the improvement.
PF-AFN~\cite{pf-afn} proposes a “teacher-tutor-student” knowledge distillation strategy and formulates it as distilling the appearance flow between the in-shop clothing image and the human body to find their accurate dense correspondences.
SDAFN~\cite{sdafn} adopts a novel deformable attention flow to achieve single-stage try-on, where the clothing warping and try-on synthesis are performed in parallel. 
HR-VTON~\cite{hr-vton} proposes a try-on condition generator as a unified module of the clothing warping and segmentation generation, which eliminates the misalignment and pixel-squeezing artifacts between the warped clothing and the segmentation map.
In addition, some methods~\cite{viton-gt, viton-gt2, ovnet} also adopt the affine transformation to assist in clothing warping.
For instance, OVNet~\cite{ovnet} learns a family of warps based on the affine transformation to describe the deformation of in-shop clothing.
VITON-GT~\cite{viton-gt2,viton-gt} uses the affine transformation to adapt the in-shop clothing to the person's pose and uses the TPS transformation to warp the clothing to fit the person's shape, respectively.
In this paper, we adopt a two-stage alignment strategy to produce warped clothing progressively.
Unlike previous methods, our clothing warping module is a combination of the affine transformation and an appearance flow, and we discard the parameters of shear mapping and only preserve the translation and scaling in the affine transformation stage, which avoids unwanted deformations that might compromise the coherence between the pre-aligned clothing and the in-shop clothing.

\section{Proposed Method}
\subsection{Overview}
Given an in-shop clothing image and a person image, our PL-VTON aims to generate the image of the person wearing the target clothing through three sub-modules, which is shown in Fig. \ref{fig:framework}.
Specifically, Progressive Clothing Warping (PCW) first takes an in-shop clothing image $C$, an in-shop clothing mask $M$, a source parsing map $P^s$ (i.e., the semantic parsing of the person image $I$), and a keypoint map $K$ (for describing the person's posture) as inputs to produce a pre-aligned clothing image $C_a$ and its corresponding mask $M_a$ through a pre-alignment network. Then a multi-scale flow predictor is subsequently adopted to obtain a pixel-level aggregated flow, which is applied to $C_a$ and $M_a$ to get a warped clothing image $C_w$ and its corresponding mask $M_w$.
The above inputs and the warped clothing image $C_w$ are fed into the next module Person Parsing Estimator (PPE) with the person image $I$ to generate a target parsing map $P^t$ (i.e., the semantic parsing prediction of the person wearing new clothing), where a non-limb target parsing map $P^t_{nl}$ is introduced as a prior parsing result in the intermediate process through the semantic combination of the warped clothing image $C_w$ and the source parsing map $P^s$.
Finally, Limb-aware Texture Fusion (LTF) takes $C_w$, $K$, $P^t$, and an occluded person image $I_{occ}$ as inputs to fuse textures of the warped clothing and the human body, where a set of limb map patches $L_p$ is adopted to refine the coarse fusion result and output the fine try-on image $I_f$.

\begin{figure*}[!t]
    \centering
    \resizebox{\linewidth}{!} {
        \includegraphics{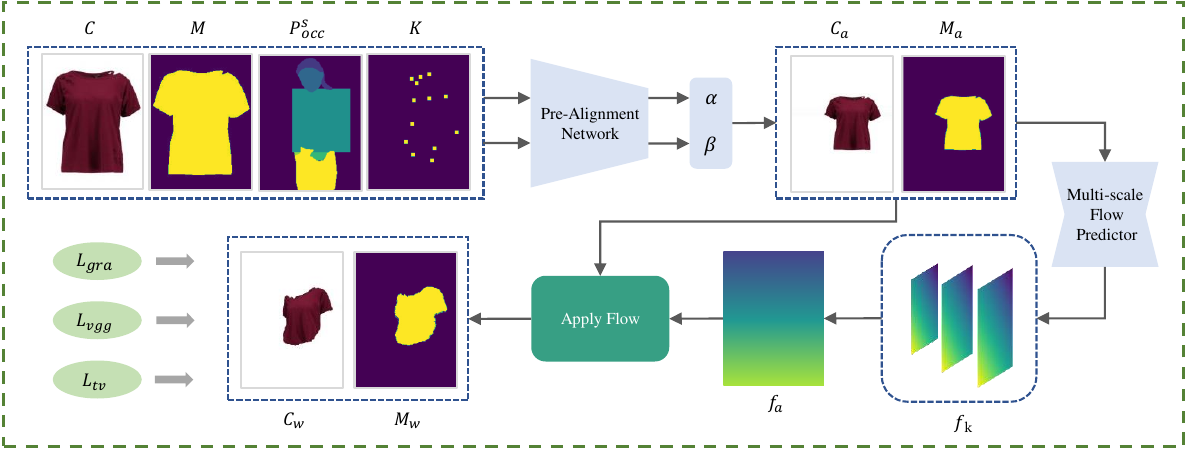}
    }
    \caption{The illustration of Progressive Clothing Warping (PCW). Given an in-shop clothing image $C$, an in-shop clothing mask $M$, an occluded source parsing map $P^s_{occ}$, and a keypoint map $K$, PCW first estimates the parameters of translation and scaling through a two-branch network, which are used to adjust the location and size of the clothing in the in-shop clothing image to get a pre-aligned clothing image $C_a$. Then a set of sub-flows are predicted and aggregated to one through the multi-scale flow predictor, which is applied to $C_a$ and $M_a$ to get a warped clothing image $C_w$ and its mask $M_w$.}
    \label{fig:frameworka}
\end{figure*}

\subsection{Progressive Clothing Warping}
To produce a fine-grained flow that warps the in-shop clothing image, we adopt Progressive Clothing Warping (PCW) with a two-stage alignment strategy, which is illustrated in Fig. \ref{fig:frameworka}. The location and size of the clothing in the in-shop clothing image are first adjusted through a pre-alignment network to locally align the human body, then a multi-scale flow predictor is used to globally estimate the geometric deformation of the clothing and get the warped clothing image $C_w$.

\subsubsection{\textbf{Clothing-agnostic Person Representation}} 
\begin{figure}[!t]
  \centering
  \resizebox{\linewidth}{!} {
    \includegraphics{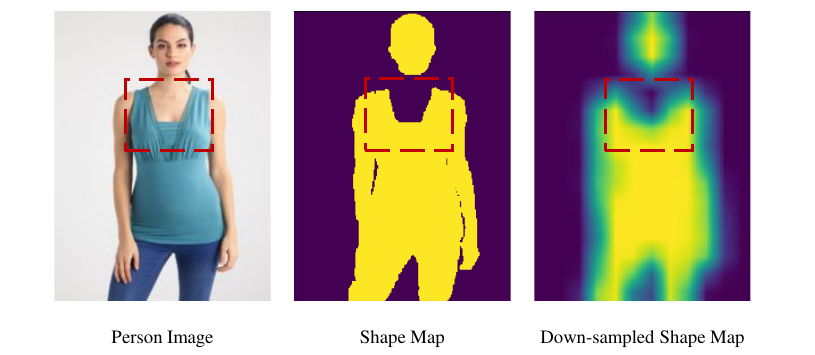}
  }
  \caption{The human shape map contains geometric information of the person's original clothing such as the collar shape, even if it has been down-sampled.}
  \label{fig:shape}
\end{figure}

Due to the difficulty of acquiring paired data of a person wearing two different clothing in a fixed posture, most existing methods adopt clothing-agnostic person representations to perform self-supervision training, which means the clothing regions in the person image are occluded before entering a network. However, these representations make the network discard some regions besides clothing and limbs such as the person's face and hair, which is negative to retain the person's other characteristics in the try-on result.
Therefore, we adopt an improved clothing-agnostic person representation to better utilize the non-clothing characteristics in the person image. We first use the circumscribed rectangle of the clothing regions in the person image as a basic mask, which is obtained by the extreme values of the clothing pixel positions in four directions. Then the boundary of the mask is modified according to the person's exposed limb regions. Finally, the source parsing map $P^s$ is used to cover the regions except for clothing and limbs of the mask. 
This mask is designed to occlude the source parsing map $P^s$ and the person image $I$ for the clothing-agnostic person representation, which is able to discard the person's original clothing while preserving the person's other characteristics.
On the other hand, it is a common practice for some methods~\cite{minar2020cp, jandial2020sievenet} to take a human shape map as one of the network inputs, which is abandoned in this paper since it contains the geometric information of the person's original clothing such as the collar shape and is not a clothing-agnostic representation strictly, even if it has been down-sampled, which is shown in Fig. \ref{fig:shape}.

\subsubsection{\textbf{Explicit Modeling in Location and Size}}
Due to the different camera views, the location and size of the clothing in the in-shop clothing image are usually inconsistent with the ones of the clothing in the person image. To handle the fine-grained clothing warping and improve the controllability of virtual try-on, we introduce a pre-alignment network to obtain the dynamic parameters that explicitly adapt the location and size of the in-shop clothing before the flow-based geometric deformation.
The pre-alignment network consists of two branches, which regress two translation parameters $\alpha_1,\alpha_2$ and two scaling parameters $\beta_1,\beta_2$, respectively. Then these parameters are arranged and combined into a matrix of size $2 \times 3$ for affine transformation, which is applied to the in-shop clothing image $C$ and the in-shop clothing mask $M$ to get the pre-aligned results. The above process is formulated as:
\begin{equation}
    C^{i,j}_a =
    \begin{bmatrix}
    \alpha_{1} & 0 & \beta_{1} \\
    0 & \alpha_{2} & \beta_{2}
    \end{bmatrix}
    C^{i,j},
\end{equation}
\begin{equation}
    M^{i,j}_a =
    \begin{bmatrix}
    \alpha_{1} & 0 & \beta_{1} \\
    0 & \alpha_{2} & \beta_{2}
    \end{bmatrix}
    M^{i,j},
\end{equation}
where $C_a$ represents the pre-aligned clothing image, $M_a$ represents the pre-aligned clothing mask, and $i,j$ is the position of the sample pixel. 
Compared with the conference version~\cite{pl-vton} of our method, in this paper, we abandon the scheme that relies on the center point of the circumscribed rectangle of clothing regions in the person image to calculate the affine transformation parameters, which is replaced by a deep neural network. This change improves the robustness of the parameter acquisition and avoids the negative impact of special human poses on the transformation parameters.

\subsubsection{\textbf{Multi-scale Flow Predictor}}
The multi-scale flow predictor consists of a 5-layer encoder based on ResNet34~\cite{he2016deep}, a convolution gated recurrent unit (ConvGRU), and a 5-layer decoder, which aims to generate an aggregated appearance flow in two steps.
In the first step, the pre-aligned clothing image $C_a \in \mathbb{R}^{3 \times H\times W}$, the keypoint map $K \in \mathbb{R}^{18 \times H\times W}$, the occluded source parsing map $P^s_{occ}\in \mathbb{R}^{7 \times H \times W}$, and the occluded person image $I_{occ}\in \mathbb{R}^{3 \times H\times W}$ are fed into the predictor, where $H \times W$ represents the resolution of input images. With the downsampling process, each encoder layer produces a set of feature maps at a specific scale, and the channel dimension of all the feature maps is uniformly adjusted to 2 by a convolution layer to get a corresponding sub-flow $f_k \in \mathbb{R}^{ 2 \times \frac{H}{6-k} \times \frac{W}{6-k}}$, where $k\in\mathbb\{1,2,...,5\}$, and the first dimension represents the pixel displacements horizontally or vertically.
The second step is to aggregate these sub-flows. Inspired by~\cite{zflow}, we use a convolution gated recurrent unit~\cite{siam2017convolutional} to perform this process, where updating and resetting operations are used to gate and combine the sub-flows through multiple non-linear weighted summations.
Specifically, in the $k$-th encoder layer, a corresponding hidden candidate state $\tilde{h}_k$ and a hidden state $h_k$ are calculated based on the current sub-flow $f_k$ and the previous hidden state $h_{k-1}$, which are formulated as:
\begin{equation}
    \tilde{h}_k = tanh(W_{fh}f_k + W_{hh} (r_{k} \odot h_{k-1})),
\end{equation}
\begin{equation}
    h_k = (1-z_k) \odot h_{k-1} + z_k \odot \tilde{h}_k,
\end{equation}
where $\odot$ is the element-wise multiplication, $r_k$ and $z_k$ are the reset and update gates, respectively, which are calculated as:
\begin{equation}
    r_k = sigmoid(W_{fr}f_k + W_{hr}h_{k-1}),
\end{equation}
\begin{equation}
    z_k = sigmoid(W_{fz}f_k + W_{hz}h_{k-1}).
\end{equation}
Note that $W_{fh}$, $W_{hh}$, $W_{fr}$, $W_{hr}$, $W_{fz}$, and $W_{hz}$ mentioned above are all the learnable weights of summations.
We regard the final hidden state as the aggregate flow $f_a$. Therefore, every pixel that the aggregated flow controls is considered from multiple scales for the fine-grained deformation.
Finally, the aggregated flow $f_a$ is applied to the pre-aligned clothing image $C_a$ and its mask $M_a$, producing the warped clothing image $C_w$ and the warped clothing mask $M_w$.

\subsubsection{\textbf{Loss}} 
\begin{figure}[!t]
    \centering
    \resizebox{\linewidth}{!} {
        \includegraphics{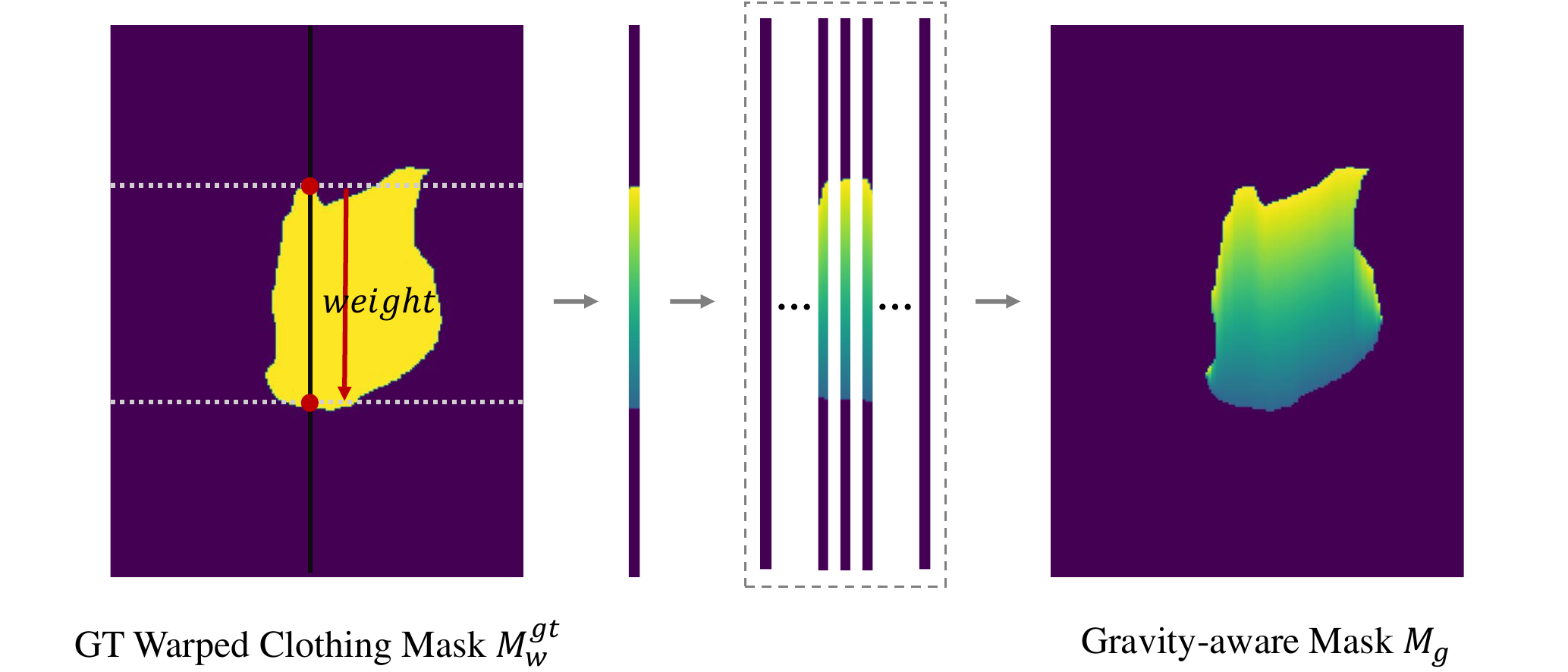}
    }
    \caption{The generation process of the gravity-aware mask $M_g$ adopted in the gravity-aware loss.}
    \label{fig:gravity}
\end{figure}

Compared with the conference version~\cite{pl-vton} of PL-VTON, we replace the mean absolute error $L_1$ with a novel gravity-aware loss $L_{gra}$. 
We consider the fit of the person wearing clothing in real scenarios, clothing regions that are close to the upper edges of the human body such as the person's shoulder fit the body more closely due to gravity, and the degree of fit decreases gradually from the top to the bottom, reaching a minimum at the hem region.
Therefore, we design a gravity-aware mask $M_g$ according to the semantic information of the GT warped clothing mask $M^{gt}_w$ (as shown in Fig. \ref{fig:gravity}, where the GT warped clothing mask that describes the person's original clothing regions is extracted from the source parsing map $P^s$). $M^{gt}_w$ is divided into $w$ vertical lines in the horizontal direction, where the value of $w$ is the same as the width of $M^{gt}_w$. Then two intersections of each vertical line and the clothing edges are found as the start and end points of the regions affected by gravity.
We weigh the regions of each vertical line in a linearly decaying way, and all the weighted vertical lines are concatenated again to form the gravity-aware mask $M_g$ finally. 
This mask is used to calculate the gravity-aware loss $L_{gra}$, which is formulated as:
\begin{equation}
    L_{gra} = \Vert (M_w - M^{gt}_w) \odot M_g \Vert_1.
\end{equation}
Furthermore, $M^{gt}_w$ is also used to get the clothing $C^{gt}_w$ in the person image, which is regarded as the ground truth of the warped clothing image $C_w$:
\begin{equation}
   C^{gt}_w = I \odot M^{gt}_w,
\end{equation}
where $\odot$ is the element-wise multiplication. $C^{gt}_w$ is used to calculate the perceptual loss $L_{vgg}$, which is formulated as:
\begin{equation}
    L_{vgg} = \sum^5_{i=1} \lambda_i \Vert \phi_i(C_{w}) - \phi_i(C^{gt}_w)\Vert_1,
\end{equation}
where $\phi_i(\cdot)$ denotes the feature maps of the $i$-th layer in the visual perception network VGG19~\cite{simonyan2014very} pre-trained on ImageNet~\cite{deng2009imagenet}, and $\lambda_i$ is the corresponding weight.
Furthermore, a total variation loss is applied to ensure the spatial smoothness of the flow $f_a$:
\begin{equation}
    L_{tv} =\sum_{i \in \Omega}\sqrt{({\rm D_x}(f_a^i))^2+({\rm D_y}(f_a^i))^2},
\end{equation}
where ${\rm D_x}(\cdot)$ and ${\rm D_y}(\cdot)$ are the horizontal and vertical difference results, respectively, and $\Omega$ is the area of $f_a$.
Finally, the whole loss function of PCW is presented as:
\begin{equation}
\label{PCWloss}
    L_{PCW} = \lambda_{gra} L_{gra} + \lambda_{vgg} L_{vgg} + \lambda_{tv} L_{tv},
\end{equation}
where $\lambda_{gra}$, $\lambda_{vgg}$, and $\lambda_{tv}$ are the weights of the loss items.

\subsection{Person Parsing Estimator}
\begin{figure}[!t]
    \centering
    \resizebox{\linewidth}{!} {
        \includegraphics{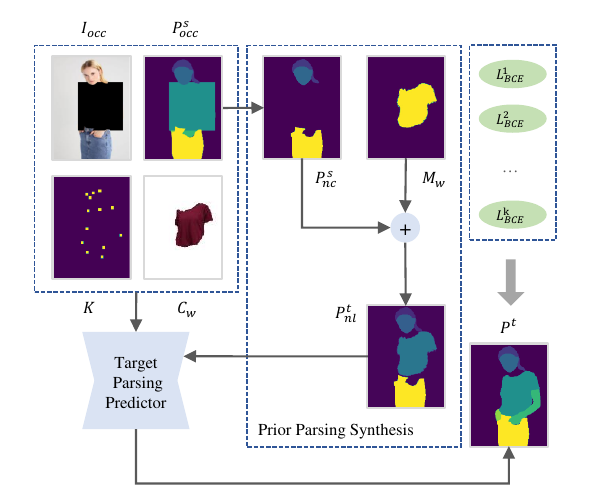}
    }
    \caption{The illustration of Person Parsing Estimator (PPE). Given an occluded person image $I_{occ}$, an occluded source parsing map $P^s_{occ}$, a keypoint map $K$, and a warped clothing image $C_w$, PPE first synthesizes a non-limb target parsing map $P^t_{nl}$ as a basic prior, then a target parsing predictor is adopted to refine the geometry of clothing regions in $P^t_{nl}$ and generate new limb regions to get the target parsing map $P^t$.}
    \label{fig:frameworkb}
\end{figure}

Some previous methods~\cite{viton, cp-vton, minar2020cp} ignore the semantic information of the human body and therefore suffer from incorrect texture generation between clothing and human skins in the try-on result, especially in the case of the transformation between long-sleeved and short-sleeved clothing.
To address this problem, we propose Person Parsing Estimator (PPE) to predict a target parsing map that represents the semantic layout of the person wearing the target clothing, which aims to provide structural constraints for the subsequent generation of the try-on result.
To further improve the accuracy of the parsing result, we introduce a non-limb target parsing map that contains the semantic information of warped clothing in the middle process of PPE, which serves as a geometric prior for the target parsing map. 
As shown in Fig. \ref{fig:frameworkb}, we first remove clothing and limb regions from the source parsing map $P^s$ and get a non-clothing source parsing map $P^s_{nc}$, then the removed regions are replaced by the warped clothing mask $M_w$ to create the non-limb target parsing map $P^t_{nl}$. Next, given the non-limb target parsing map $P^t_{nl}$, the occluded person image $I_{occ}$, the occluded source parsing map $P^s_{occ}$, the keypoint map $K$, and the warped clothing image $C_w$, a target parsing predictor is adopted to refine the geometry of clothing regions in $P^t_{nl}$ and generate new limb regions to get the target parsing map $P^t$.
The predictor is a deep neural network consisting of an encoder based on ResNet34~\cite{he2016deep} to extract semantic features from the source information in inputs and a decoder to output the target parsing result. We add an additional squeeze-and-excitation (SE) block~\cite{hu2018squeeze} after each convolutional layer to assign the weight to each channel item of inputs and feature maps, which dynamically adapts their effects on the final generated parsing map.
To train the proposed Person Parsing Estimator, we apply a weighted cross-entropy loss between the source parsing map $P^s$ and the target parsing map $P^t$, which is formulated as:
\begin{equation}
\label{PPEloss}
    L_{PPE} =-\frac{1}{n}\sum^{n}_{i=0}\sum^{c}_{j=0}{w_{j}}{P^{s}_{i,j}}log(P^{t}_{i,j}),
\end{equation}
where $n$ is the number of samples, $c$ is the number of channels of $P^s$ and $P^t$, and $w_j$ is the weight for the $j$-th class channel.
Note that we increase the weights of the clothing class and the limb class to better prevent the pixels of skins from bleeding into other regions. 

\subsection{Limb-aware Texture Fusion}
\begin{figure}[!t]
    \centering
    \resizebox{\linewidth}{!} {
        \includegraphics{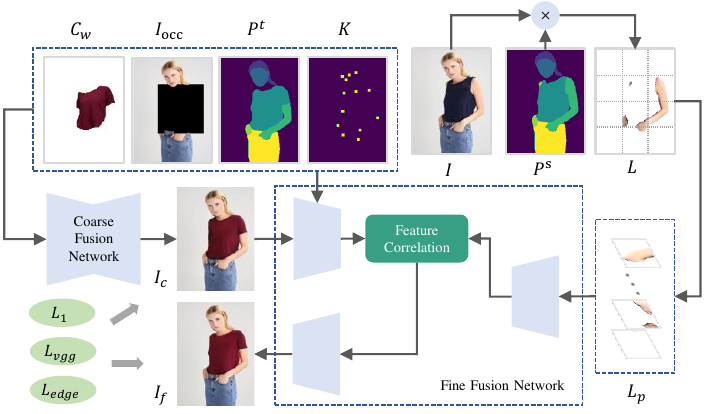}
    }
    \caption{The illustration of Limb-aware Texture Fusion (LTF), which is a coarse-to-fine generative module. Given a warped clothing image $C_w$, an occluded person image $I_{occ}$, a target parsing map $P^t$, and a keypoint map $K$, the coarse fusion network predicts the coarse-level try-on result $I_c$, which focuses on the global generation. To make the limb regions in the result more realistic, we adopt limb map patches $L_p$ in the fine fusion network, which focuses on the texture generation in the local limb regions and assists in generating the fine try-on result $I_f$.}
    \label{fig:frameworkc}
\end{figure}

Existing methods usually generate limb regions with the wrong shape and textures in the try-on result due to the discarding of limb information in the person image, which is the necessary auxiliary reference for detail generation.
Therefore, we adopt Limb-aware Texture Fusion (LTF) with the guidance of source limb textures to assist in producing realistic limb details in the try-on result.
As shown in Fig. \ref{fig:frameworkc}, LTF consists of two stages: (1) predicting the coarse try-on result $I_{c}$ and (2) refining $I_{c}$ under the limb-aware guidance to get the fine try-on result $I_{f}$.

\subsubsection{\textbf{Coarse Try-on}}
The first stage of LTF focuses on fusing the warped clothing image with the person image roughly to generate the appearance in clothing regions and other non-limb regions where textures are easily transferred, which focuses on the global generation.
As shown in Fig. \ref{fig:frameworkc}, given a warped clothing image $C_w$, an occluded person image $I_{occ}$, a target parsing map $P^t$, and a keypoint map $K$, a coarse fusion network based on ResNet34 produces the coarse try-on result $I_c$, where $C_w$ and $I_{occ}$ offer textures of target clothing and other non-limb regions that need to be preserved, respectively, $K$ and $P^t$ provide structural constraints on the generated human body to ensure the person's posture is consistent with the original image and prevent textures from being generated in wrong regions.
Although limb textures are also generated in this stage, there are some inaccurate details since plenty of related information that can be referenced in the person image is occluded, which is optimized in the next stage with the limb-aware guidance.

\begin{figure*}[!t]
  \centering
  \resizebox{\linewidth}{!} {
    \includegraphics{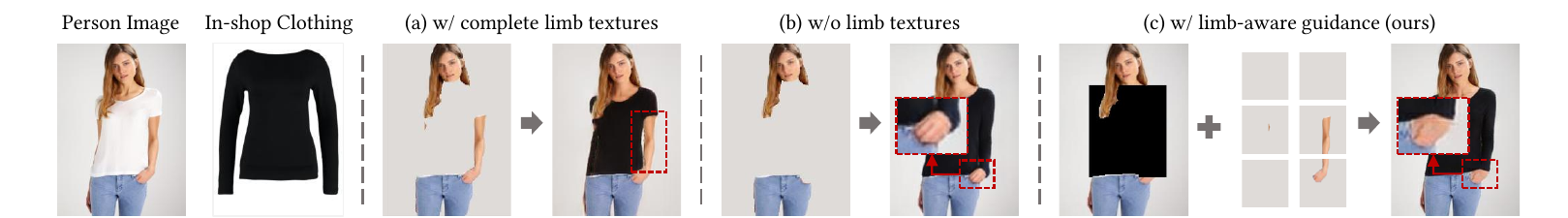}
  }
  \caption{The effect of the limb textures in inputs on the try-on result. (a) The wrong try-on result only transfers the incomplete clothing that does not match its original category since complete limb regions are preserved in inputs and the rough outline of the person's original clothing is still reflected. (b) Some artifacts occur in the try-on result because there is no assistance from source limb textures. (c) Limb-aware guidance can fuse limb textures with the try-on result while hiding the geometric information of the person's original clothing to preserve realistic limb details more reasonably.}
  \label{fig:limb}
\end{figure*}

\subsubsection{\textbf{Limb-aware Refinement}}
We train our network in a self-supervised way and try to make the virtual try-on result become exactly the same as the person image during the training process, which is easily achieved by inputting a large number of source textures of the person image.
However, it is not what we expected that this still occurs in the testing set, which means that too much information from the source space causes the network to converge to a local optimum in the training set. 
This problem is especially serious when the clothing category changes during virtual try-on. For example, we try to turn a person's short-sleeved clothing into long-sleeved clothing, it usually fails when the network receives complete limb textures from the person image, which is shown in Fig. \ref{fig:limb} (a), the wrong result only transfers the incomplete clothing that does not match its original category.
We argue the reason for this is that when we obtain the source information as a synthetic reference by removing clothing regions along the edges in the person image, the rough outline of the original clothing is still reflected through non-clothing regions, which does not keep clothing-agnostic strictly. 
It seems that we need to discard the limb textures to further hide the geometric appearance of the person's original clothing in inputs.
However, when we remove both clothing regions and limb regions from the person image and use the result as the network input, there are obvious artifacts in the limb regions of the try-on result (e.g., the person's left-hand in Fig. \ref{fig:limb} (b)), which shows that the source limb textures are necessary as auxiliary information and they have the ability to preserve realistic limb details in the try-on result.
In order to properly utilize the limb textures of the person image while keeping clothing-agnostic for training, we adopt limb-aware guidance based on a set of limb map patches, which completely hides the geometric information of the person's original clothing while fusing limb textures with the result (as shown in Fig. \ref{fig:limb} (c)).
The limb-aware guidance is introduced in the second stage of LTF to refine the coarse result.  
Specifically, a limb map $L$ that describes the exposed limb regions is first obtained by the masking operation based on the target parsing map $P^t$ and the person image $I$.
Then, we divide $L$ into patches $L_p \in \mathbb{R}^{\frac{H}{s} \times \frac{W}{s}}$, where $s$ represents the patch scale.
Finally, all the $L_p$ items are concatenated in the channel dimension and fed into the fine fusion network with $K$, $I_{occ}$, $P^t$, and $I_c$ to generate the fine try-on result $I_f$.
The fine fusion network is also a U-shaped structure with a ResNet34-based encoder and a 5-layer decoder, where a feature correlation block is used to combine the features of patched limbs and others.

\begin{figure*}[!t]
    \centering
    \includegraphics[width=1\textwidth]{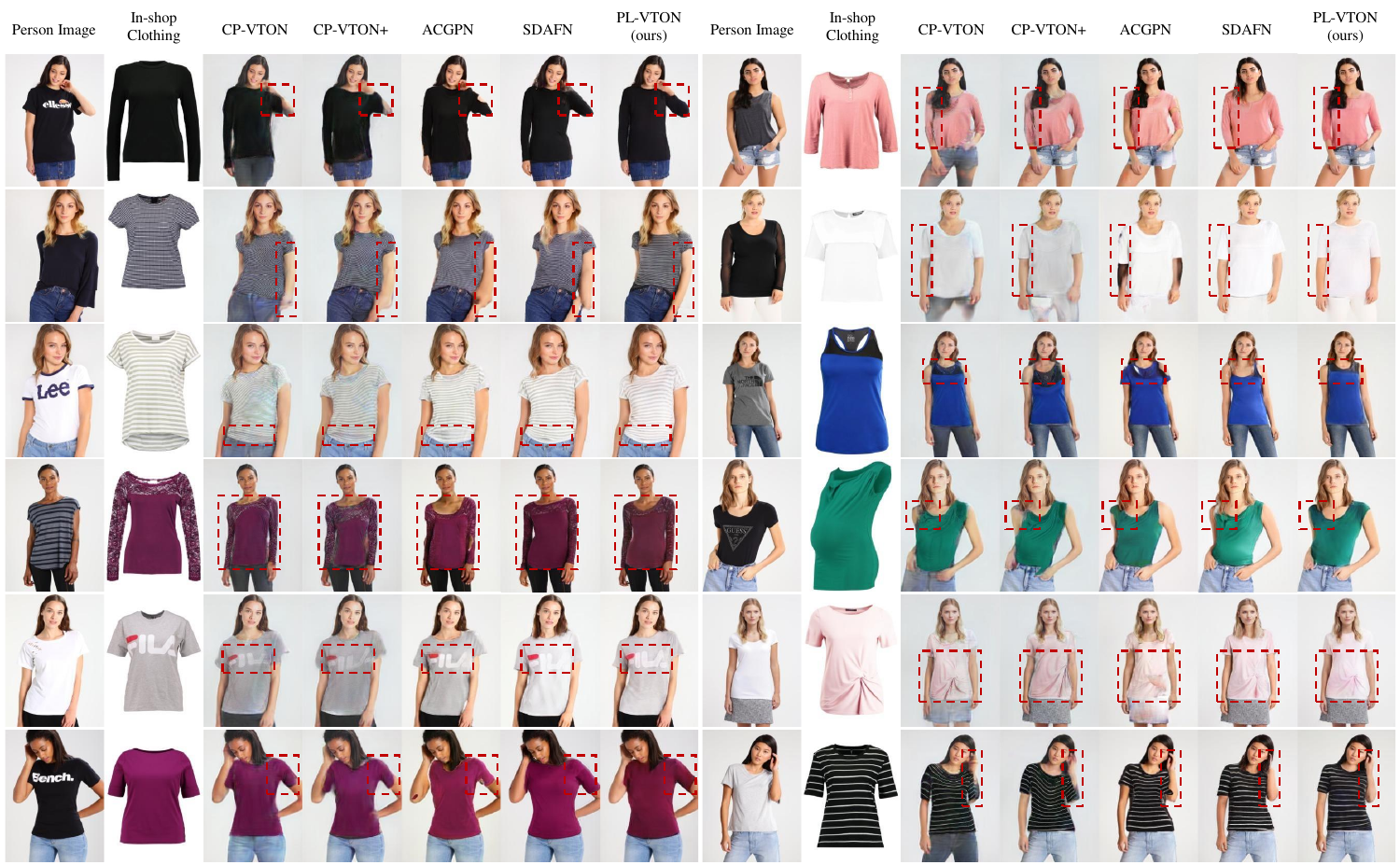}
    \caption{Visual comparisons of five different methods. The results show that PL-VTON works better than the state-of-the-art methods in the scenes of the transformation between long and short sleeves (the first row and the second row), the identification of collars and hems (the third row), unusual in-shop clothing images (the fourth row), various clothing textures (the fifth row), and complex human poses (the last row).}
    \label{fig:result} 
\end{figure*}

\subsubsection{\textbf{Loss}}
The total loss of LTF is defined as $L_{LTF}=L_{c} + L_{f}$, where $L_c$ is the loss of the coarse stage and $L_f$ is the loss of the fine stage. Both of them contain the mean absolute error, the perceptual loss, and the edge loss. Note that we use the edge loss based on Sobel filters to correct the horizontal and vertical gradients of the result, which improves the smoothness of the generated textures. $L_c$ is defined as: 
\begin{equation}
\label{LTFloss1}
\begin{split}
        L_{c} &= \lambda_{img}\Vert I_c - I \Vert_1 
        + \lambda_p (\sum^5_{i=1} \lambda_i \Vert \phi_i(I_c) - \phi_i(I)\Vert_1) \\
        &+ \lambda_{edge}\Vert\psi(I_c)-\psi(I)\Vert_1,
\end{split}
\end{equation}
where $\psi(\cdot)$ denotes the gradients extracted by Sobel filters.
Similarly, $L_f$ is defined as:
\begin{equation}
\label{LTFloss2}
\begin{split}
    L_{f} &= \lambda_{img}\Vert I_f - I \Vert_1 
    + \lambda_p (\sum^5_{i=1} \lambda_i \Vert \phi_i(I_f) - \phi_i(I)\Vert_1) \\
    &+ \lambda_{edge}\Vert\psi(I_f)-\psi(I)\Vert_1,
\end{split}    
\end{equation}
where $\lambda_{img}$, $\lambda_{p}$, and $\lambda_{edge}$ are the weights of the loss items.

\section{Experiment}
In this section, we first describe the datasets we used and show the implementation details, which involve training details and hyperparameter settings. Then, we compare our PL-VTON with the state-of-the-art methods qualitatively and quantitatively. Next, we conduct the comparison between PL-VTON in this paper and its conference version~\cite{pl-vton}. Finally, we perform five ablation studies to analyze the effectiveness of each proposed contribution in our work.

\subsection{Dataset}
We conduct experiments on the VITON~\cite{viton} dataset mainly, which contains 19K image pairs and each pair consists of a person image and a corresponding in-shop clothing image with the resolution of 256 $\times$ 192. After removing the mislabeled data, 16K image pairs are remaining, 14K of them are used for training, and the others are used for testing.
Besides, we present the additional try-on results of our method on the MPV~\cite{mpv}, VITON-HD~\cite{viton-hd}, and Dress Code~\cite{dresscode} datasets (resized) to further demonstrate the effectiveness and generalization of our method.

\subsection{Implementation Details}
We implement PL-VTON using PyTorch on one NVIDIA RTX 3090 GPU, where three sub-modules are trained independently. Specifically, we first train PCW for 60K steps, where the loss weights $\lambda_{gra}=1$, $\lambda_{vgg}=8$, and $\lambda_{tv}=0.1$. Then we train PPE for 80K steps, where $w_0=w_1=w_2=w_6=1$ and $w_3=w_4=w_5=3$. Finally, we train LTF for 80K steps, where $\lambda_{img}=1$, $\lambda_p=2$, and $\lambda_{edge}=0.4$. Besides, the patch scale $s$ in LTF is set to 8. Note that all the loss weights and the patch scale are determined by controlled experiments. All the modules are trained with a batch size of 4 and optimized by Adam~\cite{kingma2014adam} with $\beta_1=0.5$ and $\beta_2=0.999$. The learning rate is fixed at 0.0001 in the first half of training and then linearly decays to zero for the remaining steps.

\subsection{Qualitative Results}
\subsubsection{\textbf{Visual Comparisons}}
As shown in Fig. \ref{fig:result}, we perform visual comparisons of CP-VTON~\cite{cp-vton}, CP-VTON+~\cite{minar2020cp}, ACGPN~\cite{acgpn}, SDAFN~\cite{sdafn}, and our PL-VTON, where important regions are highlighted with red boxes.
In the first row, we compare the ability of these methods to fit long-sleeved clothing to a person wearing short-sleeved clothing, which focuses on examining whether the sleeves match the original limb regions accurately.
In the second row, we compare the ability of these methods to fit short-sleeved clothing to a person wearing long-sleeved clothing. In this case, more original skin regions of the person are occluded by sleeves, it is challenging to synthesize new limb textures in these regions properly.
In the third row, we aim to evaluate the ability of these methods to identify other clothing structures. Not only the front regions of clothing but also the collar or hem regions appear in some in-shop clothing images, which are usually mistaken by most existing methods.
The fourth row shows the performance of these methods when dealing with unusual in-shop clothing, which is extremely challenging in virtual try-on. For example, clothing laces or side camera views are more likely to increase the difficulty in analyzing clothing geometry. 
Results in the fifth row focus on clothing texture transfer. We select the in-shop clothing images that contain a wide range of obvious textures so the finesse and stability of clothing warping can be compared more intuitively.
In the last row, we aim to evaluate the performance of these methods to handle complex human poses. Most existing methods are difficult to identify the edges between clothing and exposed limbs under this circumstance, which leads to texture bleeding in these regions. In contrast, PL-VTON can handle this problem better.
The above qualitative experiments show that PL-VTON can better handle most virtual try-on tasks with fine-grained clothing warping and realistic limb details.

\begin{table}[!t]
\centering
\setlength{\tabcolsep}{4mm}{
\caption{Quantitative evaluation results of six compared methods and ablation study results of our PL-VTON, where PL-VTON$^{\ast}$ is PL-VTON without the pre-alignment network, PL-VTON$^{\star}$ is PL-VTON that replaces the gravity-aware loss $L_{gra}$ with the mean square error $L_1$, PL-VTON$^{\circ}$ is PL-VTON without the non-limb target parsing map, and PL-VTON$^{\diamond}$ is PL-VTON without the limb-aware guidance.}
\resizebox{\linewidth}{!} {
\begin{tabular}{l|ccc}
\hline
Method                          & FID$\downarrow$   & SSIM$\uparrow$   & PSNR$\uparrow$   \\  \hline
CP-VTON~\cite{cp-vton}   & 19.88             & 0.7891           & 21.10            \\ 
CP-VTON+~\cite{minar2020cp}     & 16.28             & 0.8191           & 21.81            \\
ACGPN~\cite{acgpn}    & 12.69             & 0.8454           & 23.12            \\ 
PF-AFN~\cite{pf-afn}      & 12.81             & 0.8552           & 22.62            \\  
DCTON~\cite{dcton}              & 12.59             & -                & -                 \\
SDAFN~\cite{sdafn}              & 10.83             & 0.8399           & 23.49            \\ \hline
PL-VTON$^{\ast}$                & 10.80             & 0.8698           & 25.86            \\ 
PL-VTON$^{\star}$               & 10.68             & 0.8715           & 26.02            \\ 
PL-VTON$^{\circ}$               & 10.71             & 0.8627           & 25.01            \\   
PL-VTON$^{\diamond}$            & 10.96             & 0.8727           & 26.04             \\   \hline
PL-VTON (ours)                  & 10.67             & 0.8731           & 26.08             \\   \hline
\end{tabular}
\label{tab:metrics}}}
\end{table}

\begin{figure}[!t]
  \centering
  \resizebox{\linewidth}{!} {
    \includegraphics{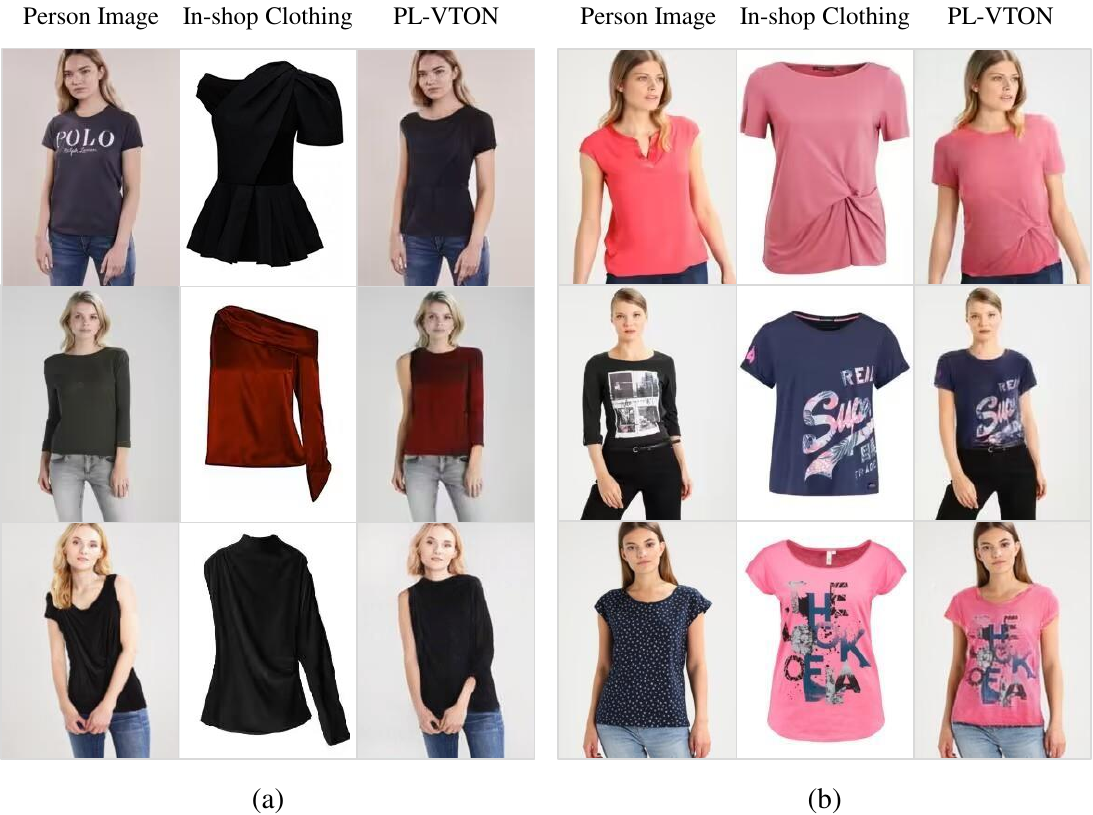}
  }
  \caption{Qualitative results of our method obtained by inputting asymmetrical clothing, where (a) focuses on the clothing with asymmetrical shapes and (b) focuses on the clothing with asymmetrical textures.}
  \label{fig:asym}
\end{figure}

\begin{table}[!t]
\centering
\caption{Quantitative comparisons between our PL-VTON and other methods on the MPV~\cite{mpv}, VITON-HD~\cite{viton-hd}, and Dress Code~\cite{dresscode} datasets.}
\resizebox{\linewidth}{!} {
\begin{tabular}{l|c|ccc}
\hline
Method                          & Dataset                                          & FID↓    & SSIM↑    & PSNR↑          \\ \hline
WUTON~\cite{wuton}              & \multirow{3}{*}{MPV}                             & 7.927   & -        & -              \\
PF-AFN~\cite{pf-afn}            &                                                  & 6.429   & -        & -              \\
PL-VTON (ours)                  &                                                  & 6.370   & 0.8551   & 23.04          \\ \hline
VITON-HD~\cite{viton-hd}        & \multicolumn{1}{c|}{\multirow{2}{*}{VITON-HD}}   & 11.44   & 0.8385   & 21.74          \\
PL-VTON (ours)                  & \multicolumn{1}{c|}{}                            & 11.35   & 0.8619   & 23.68           \\ \hline
Morelli et al.~\cite{dresscode} & \multicolumn{1}{c|}{\multirow{2}{*}{Dress Code}} & 19.63   & 0.8760    & -              \\
PL-VTON (ours)                  & \multicolumn{1}{c|}{}                            & 11.85   & 0.8780    & 23.91           \\ \hline
\end{tabular}
\label{tab:dataset metric}}
\end{table}

\begin{table}[!t]
\centering
\setlength{\tabcolsep}{4mm}{
\caption{User study results of baseline methods and our proposed PL-VTON. The results in the second column represent the percentage of the compared methods considered better than our PL-VTON, while the results in the third column represent the percentage of our PL-VTON considered better than the compared method.}
\resizebox{\linewidth}{!} {
\begin{tabular}{l|cc}
\hline
Method                             & Prefer Baseline     & Prefer PL-VTON     \\  \hline
CP-VTON~\cite{cp-vton}             & 10.94\%             & 89.06\%            \\ 
CP-VTON+~\cite{minar2020cp}        & 9.60\%              & 90.40\%            \\
ACGPN~\cite{acgpn}                 & 29.38\%             & 70.62\%            \\ 
PF-AFN~\cite{pf-afn}               & 31.90\%             & 68.10\%            \\
DCTON~\cite{dcton}                 & 26.68\%             & 73.32\%            \\ 
SDAFN~\cite{sdafn}                 & 37.88\%             & 62.12\%            \\  \hline
\end{tabular}
\label{tab:user_study}}}
\end{table}

\subsubsection{\textbf{Robustness Experiments}}
\begin{figure}[!t]
  \centering
  \resizebox{\linewidth}{!} {
    \includegraphics{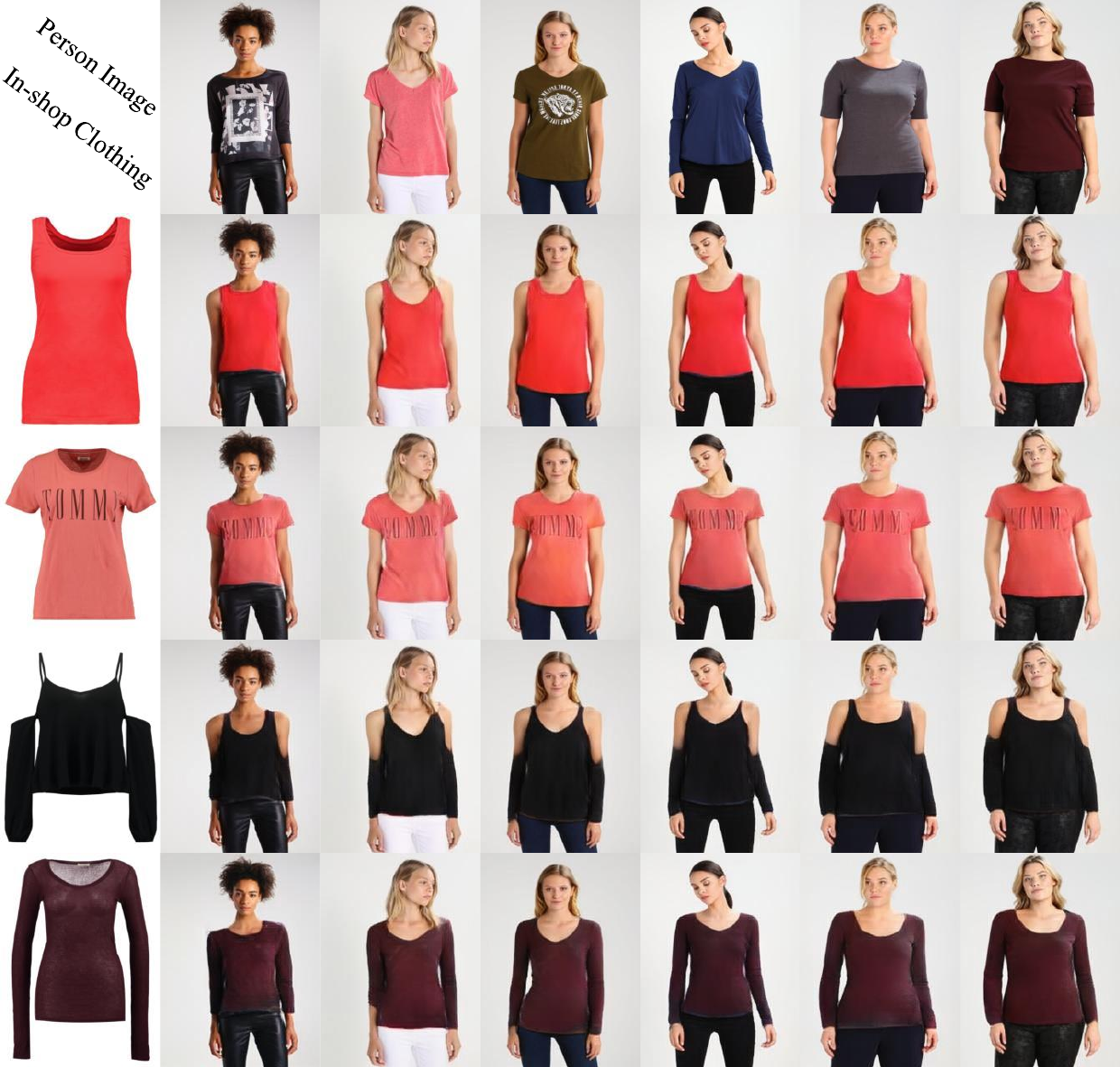}
  }
  \caption{Qualitative results of our method obtained by inputting person images with different body shapes.}
  \label{fig:body}
\end{figure}

\begin{figure*}[!t]
  \centering
  \resizebox{\linewidth}{!} {
    \includegraphics{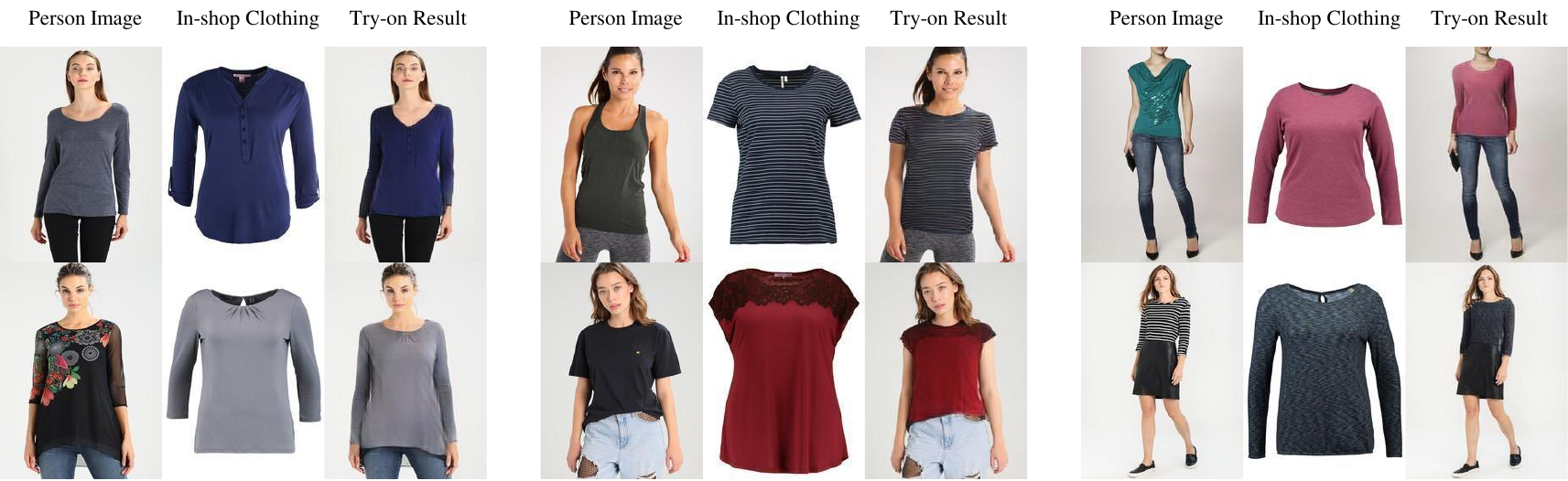}
  }
  \caption{Qualitative results of our proposed method on the MPV~\cite{mpv} dataset, which involves more various clothing styles and different person images.}
  \label{fig:mpv}
\end{figure*}

\begin{figure*}[!t]
  \centering
  \resizebox{\linewidth}{!} {
    \includegraphics{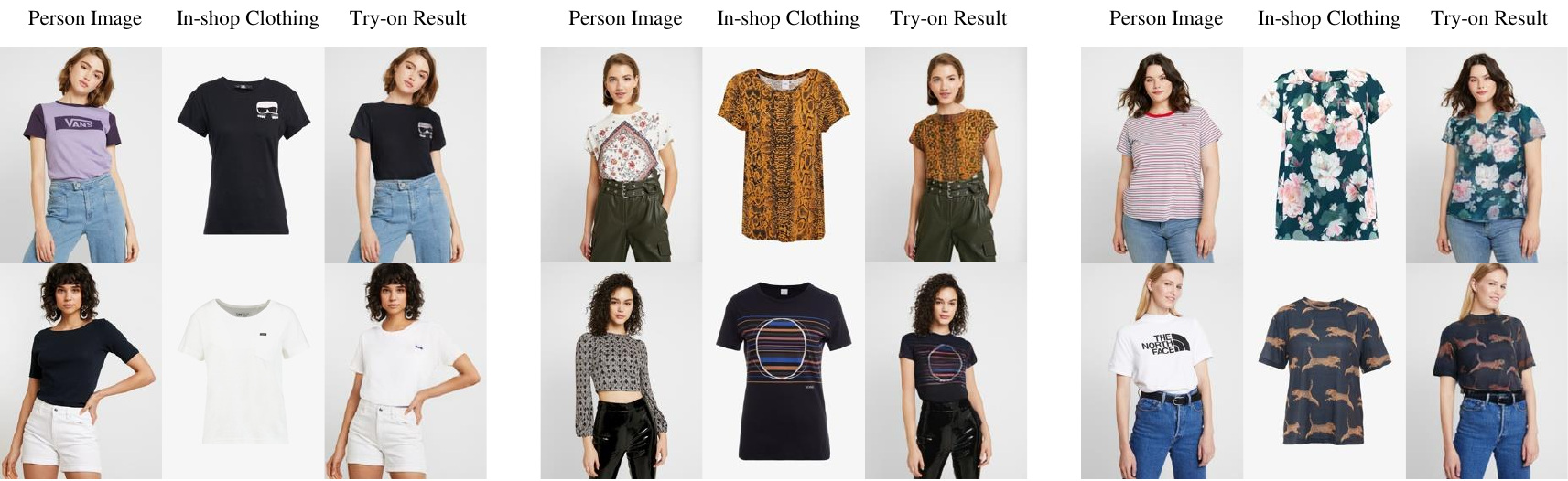}
  }
  \caption{Qualitative results of our method on the VITON-HD~\cite{viton-hd} dataset, which involves more various clothing styles and different person images.}
  \label{fig:viton-hd}
\end{figure*}

\begin{figure*}[!t]
  \centering
  \resizebox{\linewidth}{!} {
    \includegraphics{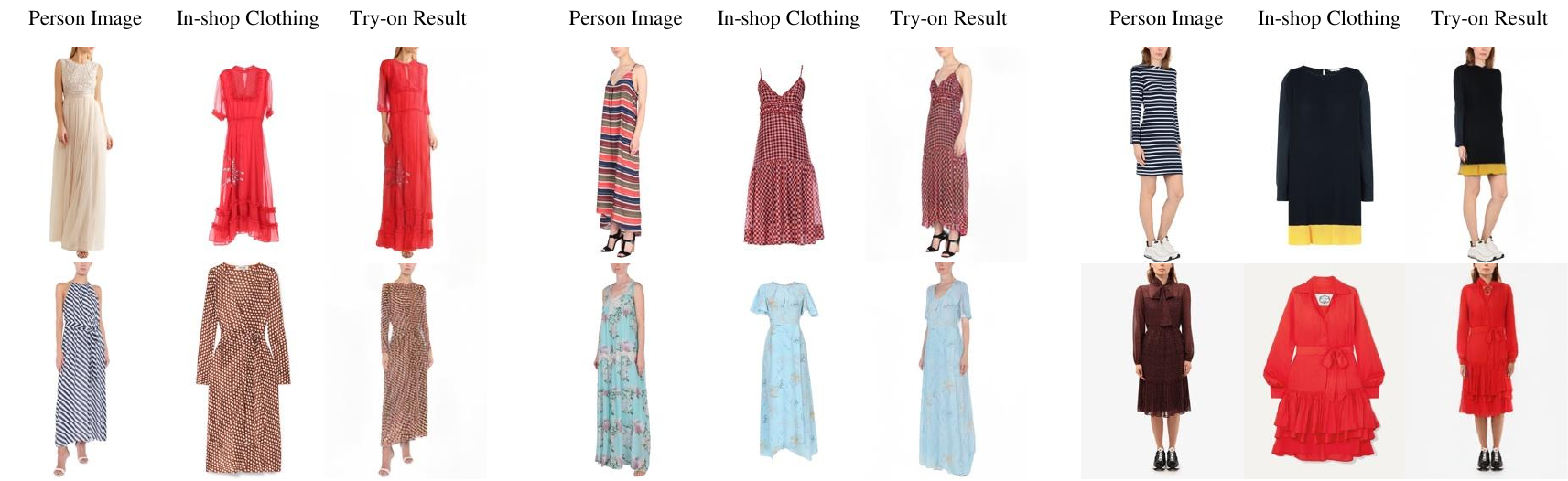}
  }
  \caption{Qualitative results of our method on the Dress Code~\cite{dresscode} dataset, which involves more various clothing styles and different person images.}
  \label{fig:dresscode}
\end{figure*}

\hspace*{\fill} \\
\noindent \textbf{Asymmetrical Clothing.} We divide asymmetrical clothing into two categories: clothing with asymmetrical shapes and clothing with asymmetrical textures. As shown in Fig. \ref{fig:asym}, our method has the ability to capture the special shapes and textures of in-shop clothing and keep these characteristics in the try-on result. \\
\noindent \textbf{Different Body Shapes.} We select several person images with different body shapes from the testing set, which are arranged for cross-matching with different in-shop clothing images. Results in Fig. \ref{fig:body} illustrate that our proposed method is capable of adaptively adjusting the clothing size to fit limbs and bodies with different sizes. \\
\noindent \textbf{Other Datasets.} As shown in Fig. \ref{fig:mpv}, Fig. \ref{fig:viton-hd}, and Fig. \ref{fig:dresscode}, we further evaluate our proposed method on the MPV~\cite{mpv}, VITON-HD~\cite{viton-hd}, and Dress Code~\cite{dresscode} datasets, which demonstrates that our method has the excellent robustness and generalization ability for various styles of in-shop clothing and different person images.

\begin{figure}[!t]
  \centering
  \resizebox{\linewidth}{!} {
    \includegraphics{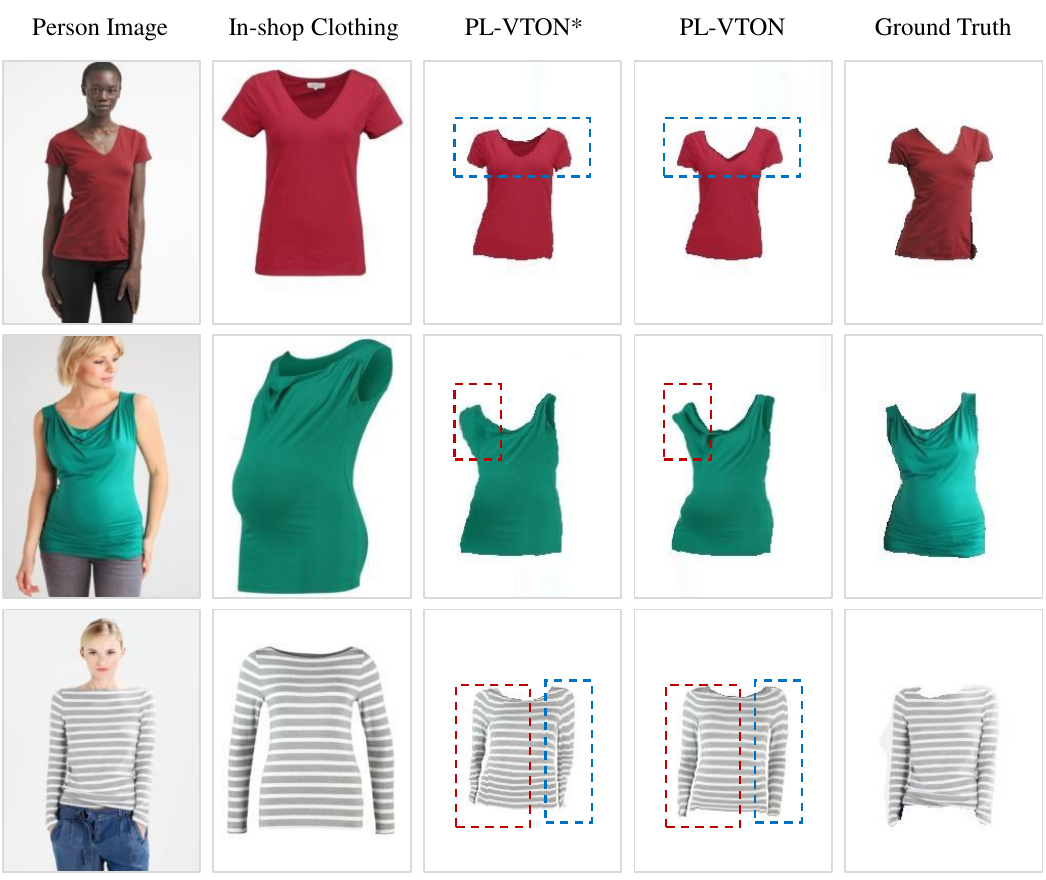}
  }
  \caption{The ablation study of the explicit modeling in location and size in Progressive Clothing Warping (PCW), where blue boxes focus on clothing geometry and red boxes focus on detailed textures. PL-VTON$^{\ast}$ is PL-VTON without the pre-alignment network.}
  \label{fig:wo_PCW}
\end{figure}

\begin{table*}[!t]
\centering
\setlength{\tabcolsep}{4mm}{
\caption{Quantitative results, computational complexity, and the user study of our method and its conference version.}
\resizebox{\linewidth}{!}{
\begin{tabular}{l|ccc|cc|c}
\hline
Method                      & FID$\downarrow$ & SSIM$\uparrow$  & PSNR$\uparrow$  & FLOPs (G) $\downarrow$  & Parameters (M) $\downarrow$  &  Human  \\  \hline
PL-VTON$^c$~\cite{pl-vton}  & 12.16           & 0.8507          & 24.73           & 119.10                  & 166.98                       &38.60\%   \\ 
PL-VTON                     & 10.67           & 0.8731          & 26.08           & 118.91                  &165.81                        & 61.40\%  \\ \hline
\end{tabular}}
\label{tab:vs}}
\end{table*}

\begin{figure*}[!t]
  \centering
  \resizebox{1\linewidth}{!} {
    \includegraphics{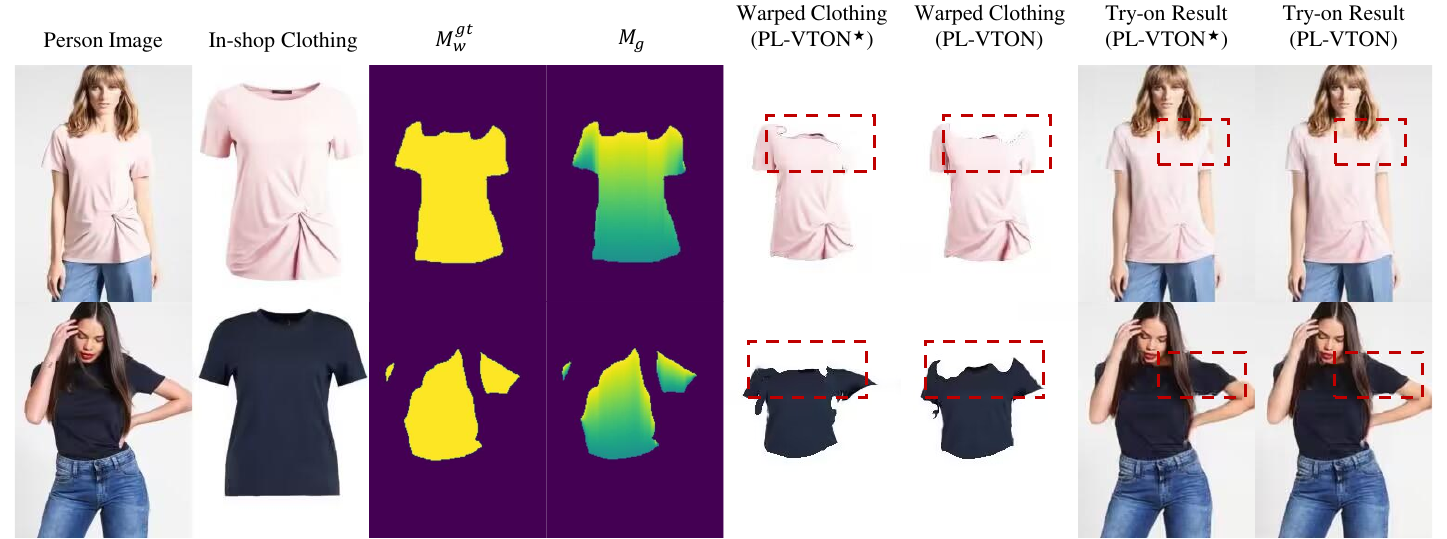}
  }
  \caption{The ablation study of the proposed gravity-aware loss in Progressive Clothing Warping (PCW), where $M^{gt}_w$ is the ground truth of the warped clothing mask, $M_g$ is the gravity-aware mask with decreasing weights from the top to the bottom, and PL-VTON$^{\star}$ is a variant of PL-VTON that replaces the gravity-aware loss $L_{gra}$ with the mean square error $L_1$.}
  \label{fig:wo_gravity}
\end{figure*}

\begin{figure}[!t]
  \centering
  \resizebox{\linewidth}{!} {
    \includegraphics{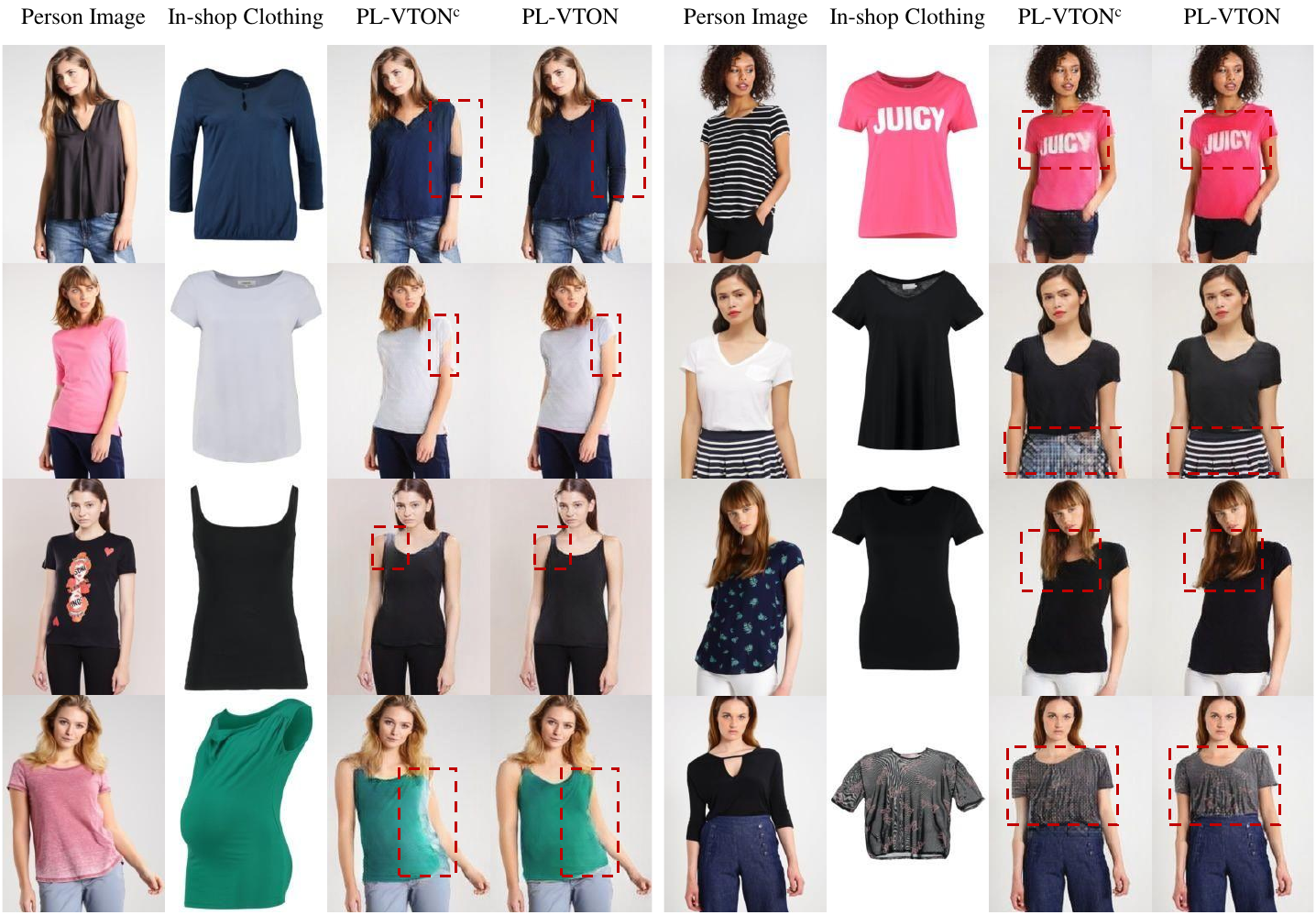}
  }
  \caption{Qualitative results of our proposed method and its conference version~\cite{pl-vton} (PL-VTON$^{c}$).}
  \label{fig:vs}
\end{figure}

\subsection{Quantitative Results}
We adopt Fréchet Inception Distance (FID)~\cite{heusel2017gans}, Structural Similarity (SSIM)~\cite{wang2004image}, and Peak Signal to Noise Ratio (PSNR)~\cite{hore2010image} to measure the performance of each method quantitatively. Better quality of results leads to lower FID, higher SSIM, and higher PSNR. Note that we do not use Inception Score (IS)~\cite{salimans2016improved} because Rosca et al.~\cite{rosca2017variational} have demonstrated that there are misleading results when IS is applied to the model trained on the datasets other than ImageNet.
Table \ref{tab:metrics} summarizes the quantitative results of CP-VTON~\cite{cp-vton}, CP-VTON+~\cite{minar2020cp}, ACGPN~\cite{acgpn}, PF-AFN~\cite{pf-afn}, DCTON~\cite{dcton}, SDAFN~\cite{sdafn}, and our PL-VTON on the VITON dataset.
Obviously, compared to the second best value (FID of 10.83, SSIM of 0.8552, and PSNR of 23.49), PL-VTON achieves better FID of 10.67, SSIM of 0.8731, and PSNR of 26.08.
On the other hand, relevant quantitative results on the MPV~\cite{mpv}, VITON-HD~\cite{viton-hd}, and Dress Code~\cite{dresscode} datasets are shown in Table \ref{tab:dataset metric}. As we can see that our PL-VTON achieves better performance than the compared methods on three datasets, which further demonstrates the robustness and generalization ability of our method.

\subsection{User Study}
A user study is conducted to further evaluate the results of our proposed method in visual effects. We perform six pairwise comparisons and each comparison involves 100 image pairs of our PL-VTON and another baseline method. 50 volunteers are asked to evaluate the quality of the two try-on results in each pair and choose the better one.
The results are summarized in Table \ref{tab:user_study}, which can be seen that PL-VTON always gives relatively better visual experiences than the compared existing methods with much higher percentages. The user study further corroborates previous evaluation results, showing that our method significantly outperforms other existing methods.

\subsection{Comparison with the previous version}
We refer to the conference version~\cite{pl-vton} of our method as PL-VTON$^{c}$ in this paper and conduct the comparisons with PL-VTON$^{c}$ in terms of the qualitative results, quantitative results, computational complexity, and user study in this section.
First, we compare our method with PL-VTON$^{c}$ on the VITON dataset qualitatively. As shown in Fig. \ref{fig:vs}, our method can effectively overcome the problems of inaccurate clothing shapes and texture artifacts occurring in the results of PL-VTON$^{c}$.
Next, comparison results of our PL-VTON and PL-VTON$^{c}$ on quantitative metrics and computational complexity are shown in Table \ref{tab:vs}. Obviously, our method achieves better metric scores while maintaining the comparable computational complexity. 
Finally, to further demonstrate the effectiveness of our improvements in this paper, we also select 50 volunteers to compare 100 groups of try-on results generated by PL-VTON and PL-VTON$^{c}$. The last column in Table \ref{tab:vs} demonstrates that our method is capable of providing better visual quality.

\begin{figure}[!t]
  \centering
  \resizebox{1\linewidth}{!} {
    \includegraphics{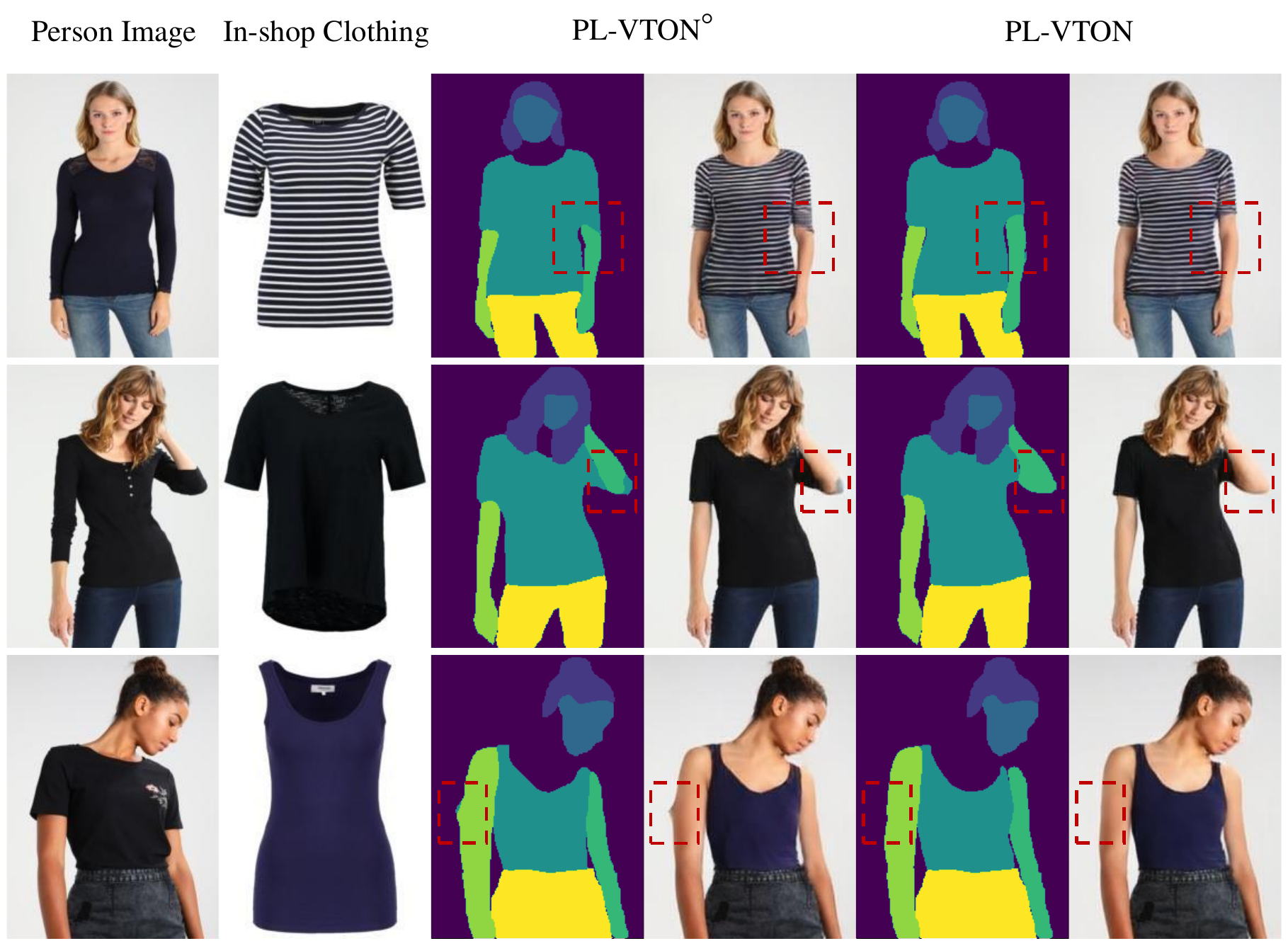}
  }
  \caption{The ablation study of the non-limb target parsing map $P^t_{nl}$ in Person Parsing Predictor (PPE), where PL-VTON$^{\circ}$ is a variant of PL-VTON without $P^t_{nl}$, the third and fifth columns are the results of the target parsing map $P^t$.}
  \label{fig:wo_PPE}
\end{figure}

\begin{figure}[!t]
  \centering
  \resizebox{1\linewidth}{!} {
    \includegraphics{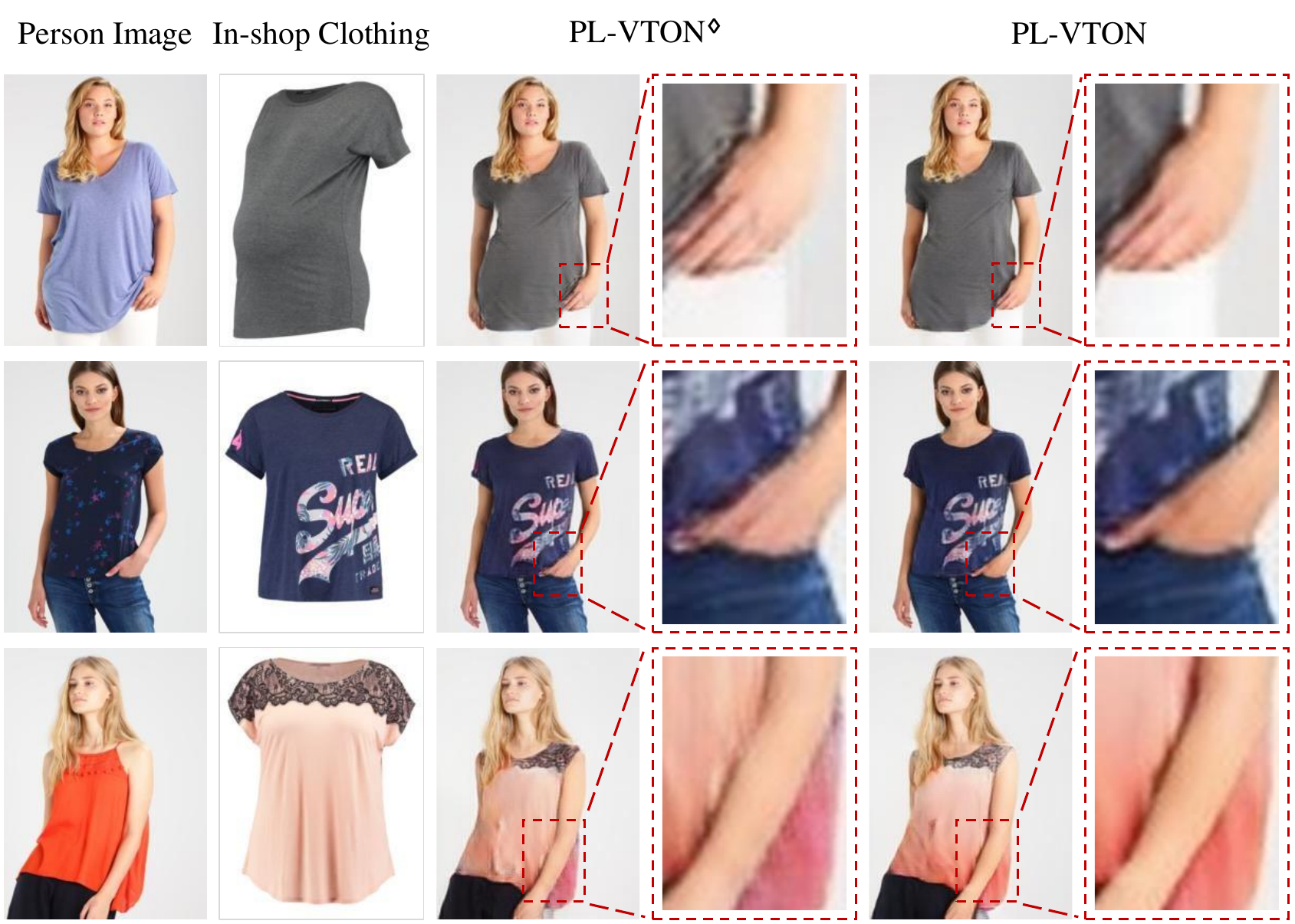}
  }
  \caption{The ablation study of the proposed limb-aware guidance in Limb-aware Texture Fusion (LTF), where PL-VTON$^{\diamond}$ is a variant of PL-VTON without the limb-aware guidance.}
  \label{fig:wo_LTF}
\end{figure}

\begin{table*}[!t]
\centering
\setlength{\tabcolsep}{4mm}{
\caption{The regions occluded by three different clothing-agnostic person representations in the person image and their impacts on the performance. PL-VTON$^{\dagger}$ does not adjust the maximum width of the occluded regions through limb regions, while PL-VTON$^{\ddagger}$ does not use the source parsing map to remove the occlusion in the regions other than clothing and limbs.}
\begin{tabular}{l|cccccc}
\hline
Method                  & occlude clothing & occlude limbs   & occlude other regions (face, hair, etc.) & FID$\downarrow$   & SSIM$\uparrow$   & PSNR$\uparrow$   \\  \hline
PL-VTON$^{\dagger}$     & $\checkmark$  & $\times$ & $\times$                               & 11.55             & 0.8611           & 24.46             \\   
PL-VTON$^{\ddagger}$    & $\checkmark$  & $part$       & $\checkmark$                           & 11.93             & 0.8520           & 24.72             \\   \hline
PL-VTON (ours)          & $\checkmark$  & $part$       & $\times$                               & 10.67             & 0.8731           & 26.08             \\   \hline
\end{tabular}
\label{tab:wo_mask}}
\end{table*}

\subsection{Ablation Studies}
We conduct five independent ablation studies on the testing set to demonstrate the role and effectiveness of each proposed contribution in our work. 
\subsubsection{\textbf{Explicit Modeling in Location and Size in PCW}}
In the first ablation study, we aim to validate the effectiveness of explicitly modeling the location and size of in-shop clothing. 
We remove the pre-alignment network in Progressive Clothing Warping, and the multi-scale flow predictor directly takes the in-shop clothing image as the input, which is denoted as PL-VTON$^{\ast}$. 
Fig. \ref{fig:wo_PCW} shows the qualitative comparisons between PL-VTON$^{\ast}$ and PL-VTON, where blue boxes focus on clothing geometry and red boxes focus on detailed textures. It can be seen that compared with PL-VTON, the results of PL-VTON$^{\ast}$ have some wrong geometric estimates around the neckline and sleeve regions, and some clothing wrinkles are missing or distorted.

\subsubsection{\textbf{Gravity-aware Loss in PCW}}
We replace the gravity-aware loss $L_{gra}$ with the mean square error $L_1$ in Eq. \ref{PCWloss} while training Progressive Clothing Warping to examine the effectiveness of $L_{gra}$, which is denoted as PL-VTON$^{\star}$. 
As shown in Fig. \ref{fig:wo_gravity}, compared with $L_1$, the novel gravity-aware loss considers the fit of the person wearing clothing in real scenarios, so it has the better ability to handle the clothing edges, where the clothing tends to interact more with the human body.
In addition, we also conduct the user study between PL-VTON and PL-VTON$^{\star}$ in Table \ref{tab:ablation_user_study} to further demonstrate the effectiveness of our gravity-aware loss.

\begin{figure}[!t]
  \centering
  \resizebox{\linewidth}{!} {
    \includegraphics{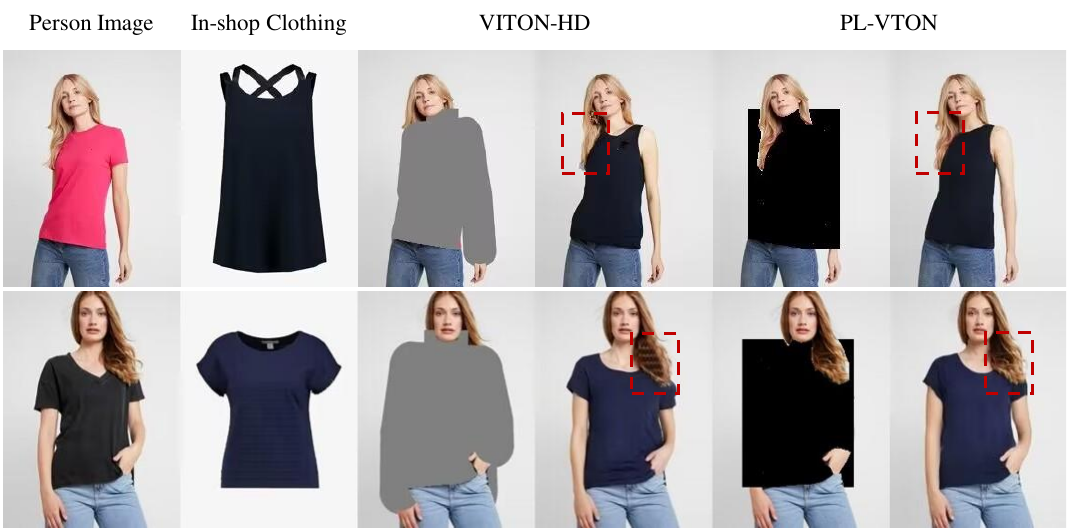}
  }
  \caption{The impact of different clothing-agnostic person representations on try-on results.}
  \label{fig:agnostic}
\end{figure}

\begin{figure*}[!t]
  \centering
  \resizebox{1\linewidth}{!} {
    \includegraphics{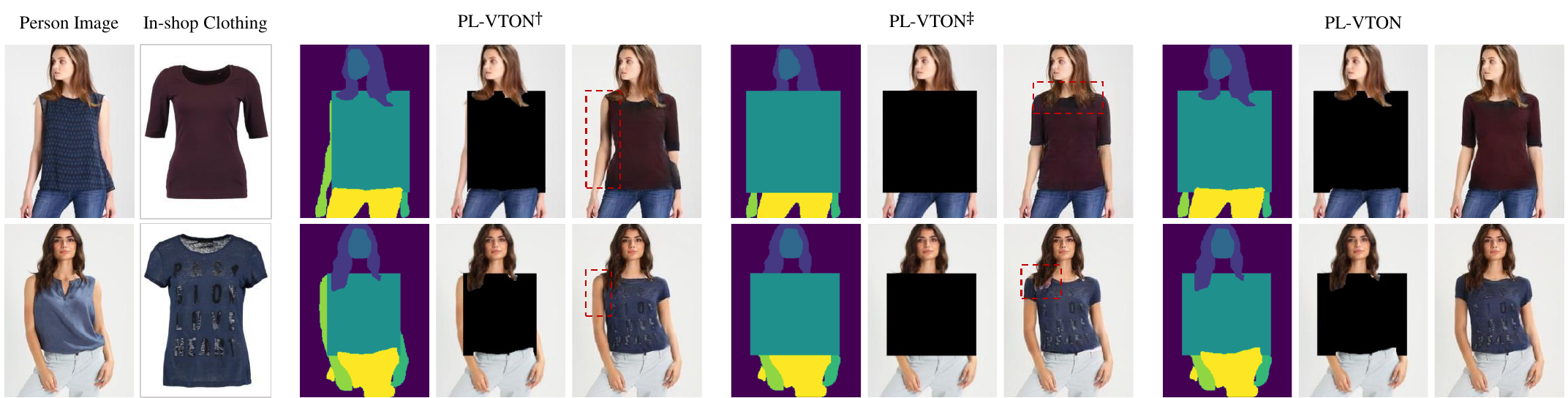}
  }
  \caption{The ablation study of our improved clothing-agnostic person representation. PL-VTON$^{\dagger}$ is a variant of PL-VTON that removes the adjustment based on limb regions to the maximum width of the occluded regions. PL-VTON$^{\ddagger}$ is a variant of PL-VTON that does not use the source parsing map to eliminate the occlusion in the regions other than clothing and limbs.}
  \label{fig:wo_mask}
\end{figure*}

\begin{table}[!t]
\centering
\caption{User studies on the gravity-aware loss and the limb-aware guidance, where PL-VTON$^{\star}$ is a variant of PL-VTON that replaces the gravity-aware loss with the $L_1$ loss, and PL-VTON$^{\diamond}$ is a variant of PL-VTON that removes the limb-aware guidance.}
\resizebox{1\linewidth}{!} {
\setlength{\tabcolsep}{4mm}{
\begin{tabular}{l|ccc|cc|c}
\hline
Method                & Prefer the variant    &  Prefer PL-VTON  \\  \hline
PL-VTON$^{\star}$     & 34.06\%               & 65.91\%            \\ 
PL-VTON$^{\diamond}$  & 35.38\%               & 64.62\%            \\ \hline
\end{tabular}
\label{tab:ablation_user_study}}}
\end{table}

\subsubsection{\textbf{Non-limb Target Parsing Map in PPE}}
We use PL-VTON$^{\circ}$ to indicate PL-VTON without the non-limb target parsing map, which means $M_w$, $P^s_{nc}$, and $P^t_{nl}$ are removed from Person Parsing Estimator and the target parsing map $P^t$ is generated only through the single target parsing predictor.
Fig. \ref{fig:wo_PPE} shows the visual comparisons of PL-VTON$^{\circ}$ and PL-VTON, where the third and fifth columns are the results of the target parsing map $P^t$. Obviously, after removing the non-limb target parsing map, some regions are incorrectly identified and generated in $P^t$ due to the lack of the prior reference, which affects the quality of the subsequent try-on result.

\subsubsection{\textbf{Limb-aware Guidance in LTF}}
We use PL-VTON$^{\diamond}$ to indicate that the limb-aware guidance is removed from Limb-aware Texture Fusion and the try-on result is generated without the assistance of limb textures in the person image. PL-VTON$^{\diamond}$ and PL-VTON are compared in Fig. \ref{fig:wo_LTF}, which shows that the limb regions generated by PL-VTON$^{\diamond}$ have some artifacts and texture bleeding while the results of full PL-VTON are more realistic and accurate.
Besides, the user study of PL-VTON and PL-VTON$^{\diamond}$ is further conducted and the result is summarized in Table \ref{tab:ablation_user_study}.

\subsubsection{\textbf{Improved Clothing-agnostic Person Representation}}
We first compare our improved clothing-agnostic person representation with two variants, which perform different occlusion operations on the person image and the source parsing map. 
In the first variant representation, we remove the adjustment based on limb regions to the maximum width of the occluded regions and denote the corresponding network as PL-VTON$^{\dagger}$. In the second variant representation, we do not use the source parsing map to eliminate the occlusion in the regions other than clothing and limbs, and denote the corresponding network as PL-VTON$^{\ddagger}$. 
As shown in Fig. \ref{fig:wo_mask}, PL-VTON$^{\dagger}$ is difficult to handle the change of sleeve length well since too much limb information in the person image is exposed to the network during the training process, which makes the transferred clothing of the try-on result incomplete and does not match its original geometric shape. As for PL-VTON$^{\ddagger}$, its occlusion operation usually occludes the characteristics that should have been preserved such as the person's face and hair, so some inconsistencies occur in the generated result.
In addition, we further compare our improved clothing-agnostic person representation with several other representations in \cite{wuton, viton-gt, dresscode}, which are produced by the segmentation mask with the dilation operation. As shown in Fig. \ref{fig:agnostic}, although both using the circumscribed rectangle and segmentation mask with the dilation operation can eliminate the negative impact of the original clothing on the try-on result, similar to PL-VTON$^{\ddagger}$, the clothing-agnostic person representations adopted in \cite{wuton, viton-gt, dresscode} do not consider that their dilation operations simultaneously occlude the regions outside clothing and limbs that need to be preserved, which results in changes in other characteristics of the person image. 
In contrast, our method masks clothing and limb regions while separately considering the regions that need to be preserved based on the source parsing map, preventing useful reference information from being lost in the input. Therefore, our clothing-agnostic person representation can achieve higher consistency in other characteristics except for clothing and limbs of the person image while ensuring the try-on effect.

The quantitative comparisons of PL-VTON$^{\ast}$, PL-VTON$^{\star}$, PL-VTON$^{\circ}$, PL-VTON$^{\diamond}$, and PL-VTON are also summarized in Table \ref{tab:metrics}, which can be seen that the performance decreases after removing the pre-alignment network, the gravity-aware loss, the non-limb target parsing map, or the limb-aware guidance, but is still better than other baseline methods basically.
Moreover, three different clothing-agnostic person representations are compared quantitatively in Table \ref{tab:wo_mask}, which shows that all the metric scores decrease significantly without our complete mask formation process, our improved clothing-agnostic person representation achieves a better balance between the source information preserved and occluded.

\section{Conclusion}
In this paper, we propose a novel virtual try-on network named PL-VTON, where three sub-modules are designed to generate high-quality try-on results, including Progressive Clothing Warping, Person Parsing Estimator, and Limb-aware Texture Fusion. On the one hand, PL-VTON explicitly models the location and size of the in-shop clothing and utilizes a two-stage alignment strategy to estimate the fine-grained clothing warping progressively. On the other hand, PL-VTON adopts limb-aware guidance to generate realistic limb details during the texture fusion between the warped clothing and the human body. Furthermore, we also design an improved clothing-agnostic person representation to 
eliminate the effect of the person's original clothing while better retaining the person's other characteristics in the try-on result. Extensive experiments demonstrate the great superiority of PL-VTON over the state-of-the-art methods for image-based virtual try-on both qualitatively and quantitatively.

\section*{Acknowledgments}
This work was supported in part by the National Natural Science Foundation of China under Grants 62272134, U21B2038, and 61976069, in part by the Taishan Scholars Program of Shandong Province under Grant tsqn201812106, and in part by the Key Project of Peng Cheng Laboratory under Grant PCL2021A07. 

\bibliographystyle{IEEEtran}
\bibliography{references}

\end{document}